\documentclass{article}

\usepackage[utf8]{inputenc} 
\usepackage[T1]{fontenc}    
\usepackage{hyperref}       
\usepackage{url}            
\usepackage{booktabs}       
\usepackage{amsfonts}       
\usepackage{nicefrac}       
\usepackage{microtype}      
\usepackage{xcolor}         

\usepackage{amsmath}
\usepackage[inline]{enumitem}
\usepackage{subcaption}
\usepackage{graphicx}
\usepackage{floatrow}
\usepackage{multirow}

\usepackage{tikz}
\usetikzlibrary{positioning,shapes,arrows,fit}
\usepackage{pgfplots}
\usepackage{mathdots}
\usepackage[multiple]{footmisc}

\usepackage[final]{neurips_2022}

\setcitestyle{numbers,square,comma}
\usepackage{float}
\floatstyle{plaintop}
\restylefloat{table}
\usepackage{xr}

\newcommand{\naive}{\mbox{Na\"{\i}ve}}
\newcommand{\seasonalnaive}{\mbox{Seasonal-na\"{\i}ve}}
\newcommand{\ets}{\mbox{ETS}}
\newcommand{\deepargaussian}{\mbox{DeepAR-Gaussian}}
\newcommand{\ctofar}{\mbox{C2FAR}}
\newcommand{\ctofarRNN}{\mbox{C2FAR-{RNN}}}
\newcommand{\ctofarone}{\ctofarRNN_{1}}
\newcommand{\ctofartwo}{\ctofarRNN_{2}}
\newcommand{\ctofarthree}{\ctofarRNN_{3}}
\newcommand{\ctofarB}{\ctofarRNN_{B}}

\newcommand{\tikzshrink}{0.88}
\newcommand{\shrinkfigthree}{0.33}
\newcommand{\shrinkfigtwo}{0.49}

\newcommand{\zhat}{\hat{z}}
\newcommand{\zvec}{\mathbf{z}}
\newcommand{\Zvec}{\mathbf{Z}}
\newcommand{\zvechat}{\mathbf{\hat{z}}}
\newcommand{\xvec}{\mathbf{x}}

\newcommand{\setb}{\mathcal{B}}
\newcommand{\setU}{\mathcal{U}}
\newcommand{\discrfunc}{d}
\newcommand{\reconfunc}{\discrfunc^{\leftarrow}}
\newcommand{\rnn}{\mbox{rnn}}

\newcommand{\elec}{\mbox{\emph{elec}}}
\newcommand{\traffic}{\mbox{\emph{traff}}}
\newcommand{\wiki}{\mbox{\emph{wiki}}}
\newcommand{\azure}{\mbox{\emph{azure}}}
\definecolor{lightlightgray}{gray}{0.9}
\tikzstyle{main} = [rectangle, draw, fill=lightlightgray, fill opacity=.6]
\tikzstyle{gblock} = [rectangle, draw, fill=lightlightgray, text centered]
\tikzstyle{wblock} = [rectangle, draw, text centered]
\tikzstyle{halo} = [rectangle, draw=none, text centered]
\tikzstyle{line} = [draw, -latex']
\tikzstyle{oval} = [ellipse, draw, fill=lightgray, node distance=2.32cm]
\tikzstyle{oval2} = [ellipse, draw, fill=lightgray, node distance=1.4cm]
\tikzstyle{dashline} = [dashed, draw, -latex']
\tikzstyle{main2} = [rectangle, draw, text width=5cm, text height=5cm, fill=lightlightgray, fill opacity=.6]
\tikzstyle{gblock2} = [rectangle, draw, fill=lightlightgray, text centered, text width=3cm]

\title{$\ctofar$\@: Coarse-to-Fine Autoregressive Networks for Precise Probabilistic Forecasting}

\author{%
Shane Bergsma \quad Timothy Zeyl \quad Javad Rahimipour Anaraki \quad Lei Guo\\
Huawei Cloud, Alkaid Lab Canada\\
\texttt{\{shane.bergsma,timothy.zeyl,javad.anaraki,leiguo\}@huawei.com}
}

\begin{document}

\maketitle

\vspace{0mm}
\begin{abstract}
  We present coarse-to-fine autoregressive networks ($\ctofar$), a
  method for modeling the probability distribution of univariate,
  numeric random variables.  $\ctofar$ generates a hierarchical,
  coarse-to-fine discretization of a variable autoregressively;
  progressively finer intervals of support are generated from a
  sequence of binned distributions, where each distribution is
  conditioned on previously-generated coarser intervals.
  Unlike prior (flat) binned distributions, $\ctofar$ can represent values
  with exponentially higher precision, for only a linear increase in
  complexity.
  We use $\ctofar$ for probabilistic forecasting via a recurrent
  neural network, thus modeling time series autoregressively in both
  space and time.
  $\ctofar$ is the first method to simultaneously handle discrete and
  continuous series of arbitrary scale and distribution shape.  This
  flexibility enables a variety of time series use cases, including
  anomaly detection, interpolation, and compression.
  $\ctofar$ achieves improvements over the state-of-the-art on several
  benchmark forecasting datasets.
\end{abstract}

\section{Introduction}\label{sec:intro}

Probabilistic forecasting is the task of estimating a joint
distribution over future values of a time series, given a sequence
of historical values.
As an important problem with many applications and an abundance of
real-world data, probabilistic forecasting has unsurprisingly come to
be dominated by deep learning methods~\cite{benidis2020neural}.
Such methods typically fit a global sequence model to a dataset of
related time series.  To forecast a given series, future values are iteratively generated from
one-step-ahead univariate output distributions.
In this paper, we propose coarse-to-fine autoregressive networks
($\ctofar$), a general method for modeling distributions of univariate
numeric data, and we show how $\ctofar$ enables improved output
distributions for probabilistic forecasting.

Such improvements are needed because time series offer challenges not
present in applications like text and image modeling.  First, dynamic
range or \emph{scale} of real-world time series can vary widely within
a single dataset~\cite{salinas2020deepar}.  The typical solution is to
normalize the data based on input historical values, but many time
series have heavy tails~\cite{ehrlich2021spliced}, meaning even after
normalization, future values must be modeled via distributions with
unbounded support.
Second, even within one dataset, time series may be discrete,
continuous, or mixed, and each of these may be distributed in
arbitrary, multi-modal ways, which vary over time.
All of this makes it difficult for practitioners to encode input
values and to select appropriate parametric forms for output
distributions.

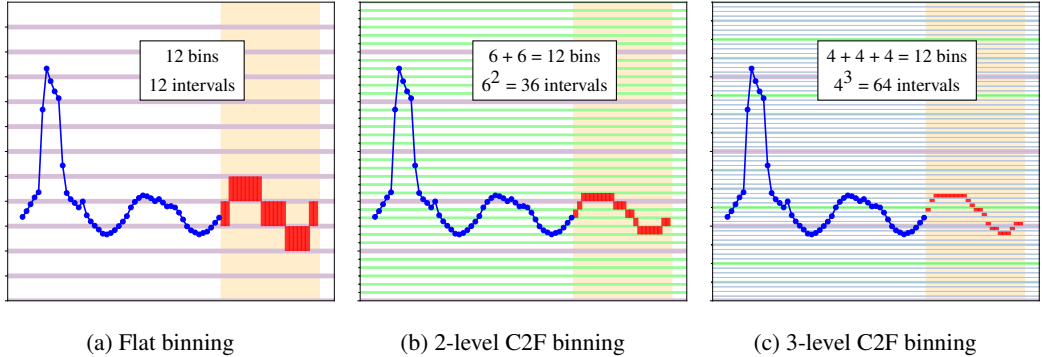
\begin{figure}
  \centering
  \vspace{-3mm}
  {\makebox[\textwidth][c]{
      \hspace{-4mm}
      \begin{subfigure}{\shrinkfigthree\textwidth}
        \captionsetup{skip=0pt}
        \begin{tikzpicture}
          \node (img1) {\includegraphics[width=\textwidth]{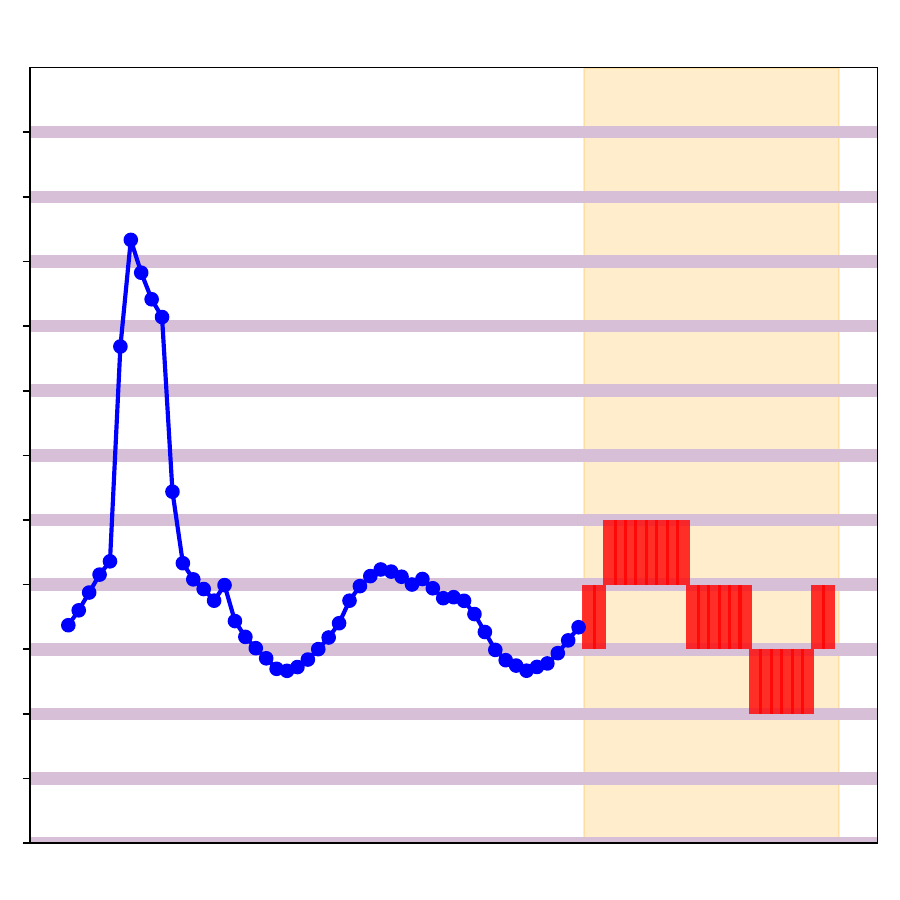}};
          \node[below of=img1, draw, align=center, fill=white, yshift=1.05cm, xshift=0.3cm, node distance=0cm, anchor=center] {\scriptsize 12 bins\\\scriptsize 12 intervals};
        \end{tikzpicture}
        \caption{Flat binning\label{fig:binningscoarse}}
      \end{subfigure}
      \hspace{-1mm}
      \begin{subfigure}{\shrinkfigthree\textwidth}
        \captionsetup{skip=0pt}
        \begin{tikzpicture}
          \node (img1) {\includegraphics[width=\textwidth]{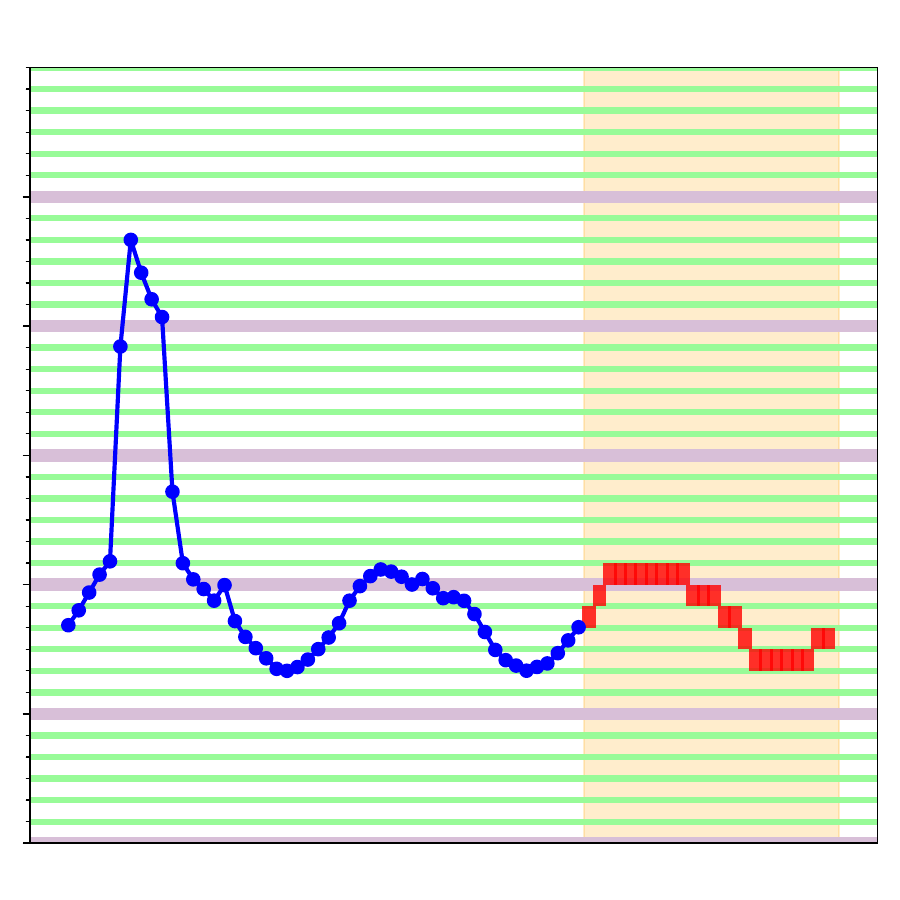}};
          \node[below of=img1, draw, align=center, fill=white, yshift=1.05cm, xshift=0.3cm, node distance=0cm, anchor=center] {\scriptsize 6 + 6 = 12 bins\\\scriptsize $\mbox{6}^{\mbox{2}}$ = 36 intervals};
        \end{tikzpicture}
        \caption{2-level C2F binning\label{fig:binningsjoint}}
      \end{subfigure}
      \hspace{-1mm}
      \begin{subfigure}{\shrinkfigthree\textwidth}
        \captionsetup{skip=0pt}
        \begin{tikzpicture}
          \node (img1) {\includegraphics[width=\textwidth]{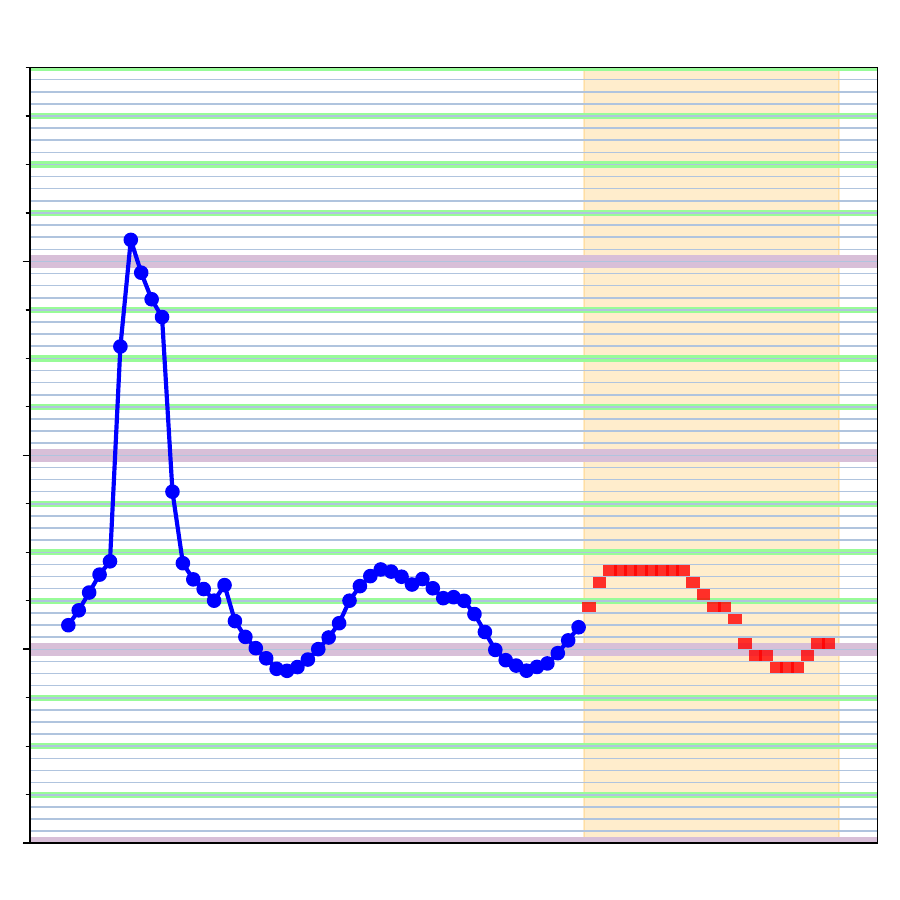}};
          \node[below of=img1, draw, align=center, fill=white, yshift=1.05cm, xshift=0.3cm, node distance=0cm, anchor=center] {\scriptsize 4 + 4 + 4 = 12 bins\\\scriptsize $\mbox{4}^{\mbox{3}}$ = 64 intervals};
        \end{tikzpicture}
        \caption{3-level C2F binning\label{fig:binningstriple}}
      \end{subfigure}
  }}
  \vspace{0mm}
  \caption{Flat vs\@. coarse-to-fine binnings for a $\traffic$ time
    series (\S\ref{sec:experiments}).  C2F discretizations specify bin
    indices from coarse-to-fine, e.g. $3,2,2$.  For the same total
    number of bins, the number of unique intervals in a C2F binning is
    much higher (and thus reconstruction error is much
    lower).\label{fig:binnings}}
\end{figure}

$\ctofar$ allows efficient modeling of any univariate numeric output
(\S\ref{sec:c2far}). $\ctofar$ extends prior work using categorical
output distributions over binned
representations~\cite{rabanser2020effectiveness,ehrlich2021spliced}.
While the precision of prior methods is limited by the number of bins
in the discretization, $\ctofar$ achieves an exponential increase in
precision through only a linear increase in the number of bin
parameters.
Like digits in the decimal number system (and unlike, say, roman
numerals or simple tally marks), $\ctofar$ represents scalar
quantities hierarchically, treating numbers as multidimensional
objects, with each successive dimension or \emph{level} representing
the original value at a progressively finer resolution
(Fig.~\ref{fig:binnings}).

$\ctofar$ proposes an autoregressive generative model for such
representations.  Levels are generated hierarchically from coarse to
fine.  Finer and finer intervals of support are generated from a
sequence of categorical (binned) distributions, each distribution
conditioned on the previously-generated coarser intervals.
The conditional distributions are parameterized by a neural network,
and the model is fit by maximizing log-likelihood of observed data.
Compared to flat binnings, $\ctofar$ models are more efficient for a
given level of precision, and better enable learning of order and
distance.

We use $\ctofar$ to model output distributions in a deep forecasting
model based on DeepAR~\cite{salinas2020deepar}
(\S\ref{sec:forecasting}).
In our experiments, $\ctofar$-based forecasting models better recover
synthetic distributions compared to recent state-of-the-art methods
(\S\ref{subsec:distribution}).
Evaluation on benchmark forecasting datasets shows that, in contrast
to prior work, discretization improves accuracy
(\S\ref{subsec:empirical}).  Multi-level $\ctofar$ models further
improve over flat binnings across all datasets, including evaluation
on a large new dataset of public cloud demand, which we release as a
paper supplement.
Code and data for C2FAR are available at
\url{https://github.com/huaweicloud/c2far_forecasting}.

\section{Background and related work}

\subsection{Probabilistic neural forecasting, use of binned output distributions}\label{subsec:forecasting}

Probabilistic forecasting architectures include
LSTMs~\cite{wen2017multi,salinas2020deepar}, temporal
convolutions~\cite{sen2019think,chen2020probabilistic} and
transformers~\cite{li2019enhancing,zhou2021informer,wu2021autoformer}.
While C2FAR can be used with any such architecture, for simplicity we
restrict our discussion in this paper to models that are
autoregressive in time~\cite{li2019enhancing,salinas2020deepar}.
While non-AR models typically output best-guess point
forecasts~\cite{cao2020spectral,sen2019think,zhou2021informer} or
specific quantiles of interest~\cite{wen2017multi,fan2019multi}, they
can be made to output C2FAR parameters at all horizons (as was done
with Gaussians in~\cite{chen2020probabilistic}).

Autoregressive networks must define step-wise input and output.  To
handle series of very different \emph{scales}, input values are often
normalized (e.g., dividing values by their
mean~\cite{li2019enhancing,salinas2020deepar,rabanser2020effectiveness});
output values are scaled back after prediction.
A density over outputs is achieved by mapping network states to
location and spread parameters of specific distributions.
Practitioners must choose a distribution ``to match the statistical
properties of the data''~\cite{salinas2020deepar}.  In practice,
Gaussian~\cite{salinas2020deepar,li2019enhancing},
Student's-t~\cite{alexandrov2020gluonts}, and Gaussian mixture
distributions~\cite{mukherjee2018armdn} have been used for real-valued
data, while a negative binomial distribution~\cite{salinas2020deepar}
has been employed for discrete.
Since output uncertainty is often not well captured by standard
parametrics, recent work has investigated more flexible outputs,
including spline quantile functions~\cite{gasthaus2019probabilistic}
and implicit quantile networks~\cite{gouttes2021probabilistic}, which
we compare to below (\S\ref{sec:experiments}).

\citet{rabanser2020effectiveness} use a categorical output
distribution over a binned representation.
Categoricals are flexible, but do not encode any underlying concept of
bin order or distance (which must be learned).
In prior work on pixel modeling, categorical distributions sometimes
work better~\cite{van2016pixel}, sometimes
worse~\cite{salimans2017pixelcnn++} than mixture densities.
In~\cite{rabanser2020effectiveness}, the benefits of binning were
mixed, and harmed DeepAR accuracy.
\citet{ehrlich2021spliced} ``splice'' a continuous Pareto distribution
into a categorical in order to model extreme values; we show how
Pareto distributions can also be used when samples are generated
autoregressively using $\ctofar$ (\S\ref{sec:c2far}).

\subsection{Challenges in forecasting mixed data, the density spike issue}\label{subsec:challenges}

Many time series are either fully discrete (e.g., counts), or
continuous but with ``clumps'' at particular values.
Dollar values are often specified to two decimal places.  Even
intrinsically continuous series are always processed or serialized to
a certain precision; for example, one version of $\elec$ has been
quantized to integers, another to six decimal places
(\S\ref{sec:experiments}).
As a cloud provider, we observe many
\emph{semicontinuous}~\cite{min2002modeling} series, with significant
probability of being either 0\% or 100\%, while otherwise being
continuous (\emph{zero-inflation} is also common in fully discrete
demand data~\cite{chapados2014effective,seeger2016bayesian}).

Modeling discrete data can be challenging.  The inherent limits of
discrete series are rarely known at prediction time, so unbounded
distributions such as negative binomials must be
used~\cite{snyder2012forecasting}.
Unfortunately, it is ``not possible'' for such distributions to output
values in a normalized domain~\cite{salinas2020deepar}, precluding
valuable scale-normalization.
But since real-valued forecasts are usually acceptable for discrete
data, practitioners often simply normalize discrete data and train
continuous models.  \citet{gouttes2021probabilistic} report better
results on $\wiki$ with a Student's-t distribution than with a
negative binomial.
\citet{rabanser2020effectiveness} normalize $\wiki$ prior to
``discretizing'' it again for their categorical.
We have also encountered practitioners with the (erroneous) belief that
normalizing makes discrete data continuous.

Unfortunately, fitting powerful continuous models to discrete/mixed
data ``can lead to arbitrarily high density values, by locating a
narrow high density spike on each of the possible discrete
values''~\cite{uria2013rnade}.
While this is a known issue in image modeling~\cite{theis2015note}, it
is apparently not well appreciated in forecasting.
Density models based on splines~\cite{gasthaus2019probabilistic} or
flows~\cite{rasul2020multi}
should therefore not be trained directly to maximize likelihood, as
this may guide parameters (e.g., spline \emph{knot} positions) solely
to enable density spikes.
A potential solution is \emph{dequantizing}, e.g., adding
\verb+Uniform[0,1]+ noise as in~\cite{rasul2020multi}.  But this
assumes:
\begin{enumerate*}[label=(\arabic*)]
 \item we know the discrete series a~priori, and
 \item the resulting loss in precision does not outweigh the benefits
   of continuous models.
\end{enumerate*}
For $\ctofar$, the ability to locate high-density spikes on discrete
values is a feature, not a bug.  $\ctofar$ does not generate
\emph{arbitrarily} narrow spikes, rather it is restricted by the
precision of the hierarchical binning.  $\ctofar$ can therefore train
for likelihood on the actual discrete data, and it can generate
realistic samples reflecting the true data distribution.

\subsection{Coarse-to-fine representations, hierarchical softmaxes}

Coarse-to-fine binnings have been used previously in machine learning.
A coarse-to-fine representation is induced by algorithms that
recursively discretize continuous-valued attributes for decision tree
classifiers~\cite{fayyad1993multi}.
On the output side, SSR-Net~\cite{yang2018ssr} classifies human age in
a coarse-to-fine (but not autoregressive) manner.
$i$SAX~\cite{shieh2008sax} uses a coarse-to-fine representation for
indexing time series.

Prior autoregressive models of discrete data can be viewed as
instances of $\ctofar$.
WaveRNN~\cite{kalchbrenner2018efficient} models 16-bit audio samples
by generating first the coarse (high) 8-bits, then the fine (low)
8-bits, conditioned on the sampled coarse.
\citet{vipperla2020bunched} found a different split between coarse
and fine bits resulted in lower cross-entropy.
These approaches can be regarded as two-level $\ctofar$ models with
classifier conditionals.
The multi-stage likelihood from~\cite{seeger2016bayesian} can be
viewed as a three-level $\ctofar$ model, with binary classification at
level~1 and~2, and Poisson regression to generate final counts.
$\ctofar$ unifies these approaches and generalizes them to unbounded,
mixed/continuous data.

Beyond numeric data, hierarchical decompositions of categorical output
distributions have been used previously in language
modeling~\cite{morin2005hierarchical,mnih2008scalable}.  As with
$\ctofar$, the key benefit is exponentially fewer computations than
when predicting words with a flat softmax.  Whether further
efficiencies such as variable-length representations (based on Huffman
coding)~\cite{mikolov2013efficient} could prove effective when
modeling continuous and ordinal data with $\ctofar$ merits further
investigation.

\section{$\ctofar$ networks}\label{sec:c2far}

The process of discretization first partitions the support of a
continuous variable into a finite number of distinct \emph{bins}.  A
value is discretized by mapping it to the index of its corresponding
bin.
With $\ctofar$, partitioning is hierarchical: the support is first
divided into a number of coarse bins, each of these coarse bins is
further divided into a number of finer bins, and so on (recall
Fig.~\ref{fig:binnings} in \S\ref{sec:intro}). $\ctofar$ thereby
discretizes a data point, $z$, into a vector of $B$ indices for each
of the $B$ coarse-to-fine levels, $\zvec=(z^1, \dots, z^B)$.  More
formally, $\ctofar$ assumes a discretization function $\discrfunc:
\mathbb{R} \to ({\setb}_1, \ldots {\setb}_B)$, where each ${\setb}_i$
has one of $|{\setb}_i|$ distinct values.
Function $\reconfunc$ is a kind of inverse of $\discrfunc$, mapping a
set of bin indices to an \emph{interval} in the original real-valued
domain, $\reconfunc: ({\setb}_1, \ldots {\setb}_B) \to
(\mathbb{R},\mathbb{R})$.

\begin{figure}
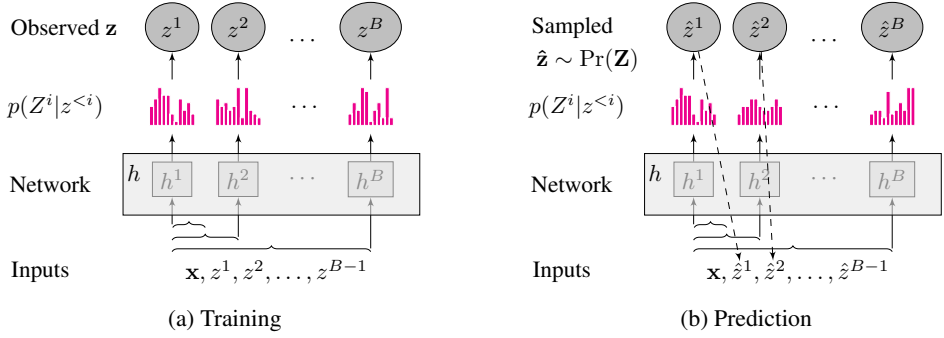

  \centering
  {\makebox[\textwidth][c]{
    \begin{subfigure}{\shrinkfigtwo\textwidth}
      \centering
      \scalebox{\tikzshrink}{
        {\input{tikz_figures/c2far_train.tex}}
      }
      \caption{Training\label{fig:c2fartrain}}
    \end{subfigure}
    \begin{subfigure}{\shrinkfigtwo\textwidth}
      \centering
      \scalebox{\tikzshrink}{
        {\input{tikz_figures/c2far_test.tex}}
      }
      \caption{Prediction\label{fig:c2fartest}}
    \end{subfigure}
  }}
  \caption{$\ctofar$ network for regression. We train to minimize NLL
    of a set of examples; log loss on each example decomposes into the
    sum of NLL of each $z^i$ under its categorical distribution.  In
    prediction, we sample from the categoricals, and autoregressively
    use the sampled values as inputs.\label{fig:c2far}}
\end{figure}

Let $\zvec = \discrfunc(z)$ be the $B$-dimensional discretization of
$z$.
Let $\zvec^{<i}$ denote the first $i-1$ bin indices, $z^1, \ldots, z^{i-1}$.
Let $\xvec$ be an optional set of features that may inform $z$'s
distribution (e.g., for mixed data, $\xvec$ could flag which output
bins contain integers).
We follow prior work in modeling \emph{multivariate} conditional
distributions $p(\zvec|\xvec)$ as autoregressive generative models via
the chain rule~\cite{frey1995does,bengio2000modeling,uria2016neural}:
\begin{align}
p(\zvec|\xvec) = \prod_{i=1}^B p(z^i|\xvec, \zvec^{<i})
\label{eqn:auto}
\end{align}
In $\ctofar$, each of the one-dimensional distributions, $p(z^i|\xvec,
\zvec^{<i})$, is a categorical over the set of bin indices at the
$i$th granularity, ${\setb}_i$.  Intuitively, the $i$th categorical is
a distribution over the support \emph{within} the generated $(i-1)$th
bin. We parameterize these distributions using neural networks with a
softmax output layer (specific architectures are discussed below), and
fit the parameters to minimize negative log-likelihood (NLL) of
training data (Fig.~\ref{fig:c2fartrain}).
C2FAR is agnostic toward the encoding of the inputs $z^{i}$; they may,
for example, be represented with 1-hot-encodings or embedding layers.
As a generative model, we synthesize new data by first sampling a
coarse bin ($\zhat^1$) given $\xvec$, then iteratively sampling finer
and finer bins given the sampled (and encoded) coarser values
(Fig.~\ref{fig:c2fartest}).

$\ctofar$ supports a variety of neural architectures: following prior
work in multivariate distribution modeling, we may use a separate
(feedforward) network for each categorical~\cite{bengio2000modeling}
(as in Fig.~\ref{fig:c2far}), or one global network with masking to
enforce the autoregressive property~\cite{germain2015made}.  Separate
networks may share parameters across the
levels~\cite{larochelle2011neural} or with classifications at the same
level across time (\S\ref{sec:forecasting}).
Having separate networks for each level means the modeling problem at
a given level need not change if we add more levels to the hierarchy;
we adopt this approach in our forecasting experiments
(\S\ref{subsec:ctofarrnn}).

While not strictly required, we use linear (evenly-spaced) binnings at
each level.  We hypothesize this enables the finer networks to better
generalize learned concepts of order and distance (i.e., regardless of
the coarse bins that they are conditioned on).
We refer to the span from the first interval to the last interval as
the \emph{extent} of the binning.  The extent, the number of levels,
and the number of bins at each level, are $\ctofar$ hyperparameters.
We also consider the extreme high/low intervals to be open-ended,
terminating at $\pm\infty$; we discuss the implications of this below.

\textbf{Level $B+1$ parametric distributions.} Let $(a,b)$ be the
interval defined by $(a,b) = \reconfunc(\zvec)$.  To complete our
generative story, we must generate a value within $(a,b)$ (rather than
choosing a single ``reconstruction value'' for each
bin as in~\cite{rabanser2020effectiveness}).
Since our extreme intervals are open-ended, we cannot interpret our
model as a piecewise uniform distribution~\cite{van2016pixel}, as
uniforms are only defined on finite intervals.
$\ctofar$ solves this by allowing different distributions to be used
depending on the interval $(a,b)$.
Essentially, we assume a final conditional in~(\ref{eqn:auto})
(implicitly at level $B+1$) that generates from a (differentiable)
parametric distribution.  We use distributions of the form:
\begin{align}
  p(Z^{B+1}|a,b)\sim\begin{cases}
    \mbox{Uniform}[a,b] & -\infty < a \mbox{ and } b < \infty \\
    \mbox{Pareto}[a,\alpha_1] & b = \infty \\
    -\mbox{Pareto}[b,\alpha_2] & a = -\infty
  \end{cases}
  \label{eqn:pareto}
\end{align}
Here $\mbox{Pareto}$ indicates a Type I Pareto distribution with fixed
scale parameter ($a$ or $b$, defined a priori from the extent of the
binning in the observed space) and dynamic shape parameter $\alpha_i$.
Parameters $\alpha_1$ and $\alpha_2$ are generated by the $h_{B+1}$
neural network in a manner analogous to how DeepAR outputs the mean
and variance parameters of a Gaussian
distribution~\cite{salinas2020deepar}.\footnote{In this way,
  ``weighted'' Pareto tails are not ``spliced'' into the distribution
  at fixed user-defined quantiles, as they are
  in~\cite{ehrlich2021spliced}. In C2FAR, Pareto tails provide the
  probability density function of values in the extreme bin,
  \emph{given the values are in the extreme bin}. Each Pareto density
  itself integrates to 1, but is dynamically ``weighted'' by the probability
  of being in the corresponding extreme bin.  Note also that if data
  does not have heavy tails, alternative distributions may be used in
  the extreme bins instead, e.g., left and right-truncated Gaussians.}
We are effectively defining a piecewise-uniform density with
Pareto-distributed tails.\footnote{By parameterizing the quantile
  function with linear splines,
  SQF-RNN~\cite{gasthaus2019probabilistic} actually enforces piecewise
  uniformity, over a \emph{finite} range.  Neural spline
  flows~\cite{durkan2019neural} require derivatives of the spline
  functions to match at knots, to avoid ``numerical issues.'' By
  parameterizing the PDF directly using classifiers, and only
  manifesting the portion needed, $\ctofar$ has none of the above
  restrictions, while being more efficient and well-behaved.}
We use this model for both continuous and discrete data.
If data are truly discrete and precision is useful, our tuning
procedure will choose more bins/levels.
If one knows a priori the true discrete support, one may instead use
discrete distributions at level $B+1$, such as negative binomials,
Poisson regression~\cite{seeger2016bayesian}, or discrete uniforms.

\textbf{Complexity.}  Consider a $\ctofar$ discretization with $B$
levels and an unvarying cardinality of $K$ bins at each level; the
original support of $z$ is effectively partitioned into $K^B$ total
intervals, but modeled using only $KB$ categorical outputs in total
($B$ softmaxes with $K$ values each).
We may regard C2FAR as a $K$-ary tree of height $B$.  At each node in
the tree, we determine which of the $K$ bins we fall into at that
level, and follow the chosen branch to the sub-tree at the next level.
We need only evaluate the softmax probabilities (and backpropagate
gradients) for the $B$ nodes on the path from the root to the leaf of
the tree.  Since the height of the tree, $B$, is logarithmic (in base
$K$) over the total number of intervals, we compute exponentially
fewer outputs with C2FAR compared to flat binnings.  The final
likelihood additionally requires computing the probability at the
$B+1$ level according to the corresponding parametric distribution
(Uniform or Pareto) as given in Eqn.~(\ref{eqn:pareto}).
In this way, full $\ctofar$ densities are never explicitly manifested
when training or predicting; for plotting (e.g.,
Fig.~\ref{fig:outputs}), we explicitly compute likelihood at tiny
increments over a predefined range.

\section{Forecasting with $\ctofar$}\label{sec:forecasting}

We now explain how $\ctofar$ can be used for $N$-step-ahead
probabilistic forecasting.  We base the forecasting framework on
DeepAR~\cite{salinas2020deepar}, which has served as the basis for
other recent improvements in forecast distribution
modeling~\cite{gasthaus2019probabilistic,gouttes2021probabilistic},
and therefore facilitates experimental comparison
(\S\ref{sec:experiments}).

\subsection{DeepAR-style probabilistic forecasting}

Let $z_t$ be the value of a time series at time $t$, and $\xvec_t$ be a
vector of time-varying features or \emph{covariates}.  Probabilistic
forecasting aims to model the conditional distribution of $N$ future
values of $z_t$ (the \emph{prediction range}) given the $T+N$
covariates, and $T$ historical values (the \emph{conditioning range}):
\begin{align}
p(z_{T+1} \ldots z_{T+N}|z_1 \ldots z_T, \xvec_1 \ldots \xvec_{T+N})
\stackrel{\text{def}}{=} p(z_{T+1:T+N}|z_{1:T}, \xvec_{1:T+N})
\label{eqn:forecast}
\end{align}

\begin{figure}
  \centering
  {\makebox[\textwidth][c]{
    \begin{subfigure}{\shrinkfigtwo\textwidth}
      \centering
      \scalebox{\tikzshrink}{
        {\begin{tikzpicture}
  \node [oval] (zt-1) {$\zvec_{t-1}$};
  \node [oval, right of=zt-1] (zt) {$\zvec_{t}$};
  \node [oval, right of=zt] (zt+1) {$\zvec_{t+1}$};
  \node [wblock, below of=zt-1] (pt-1) {$p(\zvec_{t-1}|\theta_{t-1})$};
  \node [wblock, below of=zt] (pt) {$p(\zvec_{t}|\theta_{t})$};
  \node [wblock, below of=zt+1] (pt+1) {$p(\zvec_{t+1}|\theta_{t+1})$};
  \node [gblock, below of=pt-1] (ht-1) {$h_{t-1}$};
  \node [gblock, below of=pt] (ht) {$h_{t}$};
  \node [gblock, below of=pt+1] (ht+1) {$h_{t+1}$};
  \node [halo, left of=ht-1] (ht-2) {$\cdots$};
  \node [halo, right of=ht+1] (ht+2) {$\cdots$};
  \node [halo, below of=ht-1] (int-1) {$\zvec_{t-2}, \xvec_{t-1}, \zvec_{t-1}^{<B}$};
  \node [halo, below of=ht] (int) {$\ \ \zvec_{t-1}, \xvec_{t}, \zvec_{t}^{<B}$};
  \node [halo, below of=ht+1] (int+1) {$\zvec_{t}, \xvec_{t+1}, \zvec_{t+1}^{<B}$};
  \path [line] (ht-2) -- (ht-1);
  \path [line] (ht-1) -- (ht);
  \path [line] (ht) -- (ht+1);
  \path [line] (ht+1) -- (ht+2);
  \path [line] (int-1) -- (ht-1);
  \path [line] (int) -- (ht);
  \path [line] (int+1) -- (ht+1);
  \path [line] (ht-1) -- (pt-1);
  \path [line] (ht) -- (pt);
  \path [line] (ht+1) -- (pt+1);
  \path [line] (pt-1) -- (zt-1);
  \path [line] (pt) -- (zt);
  \path [line] (pt+1) -- (zt+1);
\end{tikzpicture}}
      }
      \caption{Training\label{fig:archzoomedouttrain}}
    \end{subfigure}
    \begin{subfigure}{\shrinkfigtwo\textwidth}
      \centering
      \scalebox{\tikzshrink}{
        {\begin{tikzpicture}
  \node [oval] (zt-1) {$\zvechat_{t-1}$};
  \node [oval, right of=zt-1] (zt) {$\zvechat_{t}$};
  \node [oval, right of=zt] (zt+1) {$\zvechat_{t+1}$};
  \node [wblock, below of=zt-1] (pt-1) {$p(\zvec_{t-1}|\theta_{t-1})$};
  \node [wblock, below of=zt] (pt) {$p(\zvec_{t}|\theta_{t})$};
  \node [wblock, below of=zt+1] (pt+1) {$p(\zvec_{t+1}|\theta_{t+1})$};
  \node [gblock, below of=pt-1] (ht-1) {$h_{t-1}$};
  \node [gblock, below of=pt] (ht) {$h_{t}$};
  \node [gblock, below of=pt+1] (ht+1) {$h_{t+1}$};
  \node [halo, left of=ht-1] (ht-2) {$\cdots$};
  \node [halo, right of=ht+1] (ht+2) {$\cdots$};
  \node [halo, below of=ht-1] (int-1) {$\zvechat_{t-2}, \xvec_{t-1}, \zvechat_{t-1}^{<B}$};
  \node [halo, below of=ht] (int) {$\ \ \zvechat_{t-1}, \xvec_{t}, \zvechat_{t}^{<B}$};
  \node [halo, below of=ht+1] (int+1) {$\zvechat_{t}, \xvec_{t+1}, \zvechat_{t+1}^{<B}$};
  \path [line] (ht-2) -- (ht-1);
  \path [line] (ht-1) -- (ht);
  \path [line] (ht) -- (ht+1);
  \path [line] (ht+1) -- (ht+2);
  \path [line] (int-1) -- (ht-1);
  \path [line] (int) -- (ht);
  \path [line] (int+1) -- (ht+1);
  \path [line] (ht-1) -- (pt-1);
  \path [line] (ht) -- (pt);
  \path [line] (ht+1) -- (pt+1);
  \path [line] (pt-1) -- (zt-1);
  \path [line] (pt) -- (zt);
  \path [line] (pt+1) -- (zt+1);
  \path [dashline] (zt-1) -- (1.63,-2.8);
  \path [dashline] (zt) -- (3.84,-2.8);
\end{tikzpicture}}
      }
      \caption{Prediction\label{fig:archzoomedouttest}}
    \end{subfigure}
  }}
  \caption{\label{fig:archzoomedout}High-level $\ctofarRNN$ for
    forecasting.  Unlike DeepAR, $\ctofarRNN$ uses discretized
    input/output vectors, $\zvec_t$, rather than scalars, and has
    additional (autoregressive) inputs $\zvec_t^{<B}$.}
\end{figure}
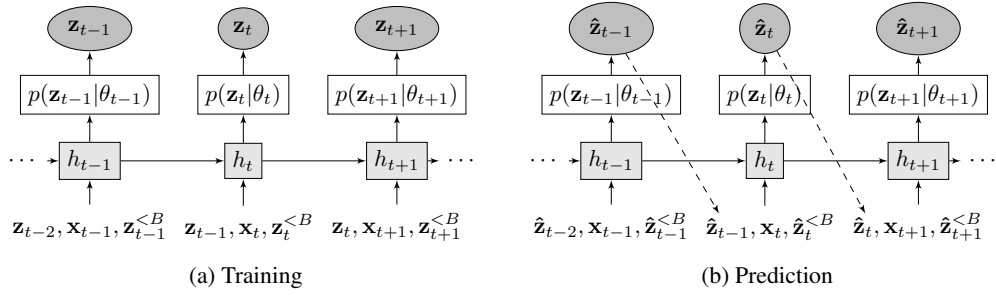

DeepAR~\cite{salinas2020deepar}
formulates this conditional distribution as an
autoregressive generative model:
\begin{align*}
p(z_{T+1:T+N}|z_{1:T}, \xvec_{1:T+N}) =& \prod_{t=T+1}^{T+N} p(z_t |z_{1:{t-1}}, \xvec_{1:{T+N}}) = \prod_{t=T+1}^{T+N} p(z_t |\theta_t=y(h_t))
\end{align*}
where $h_t = \rnn(h_{t-1}, z_{t-1}, x_{t})$ is the output of an
LSTM~\cite{hochreiter1997long} recurrent neural network,
and $y(\cdot)$ is a function that maps the output of the LSTM to the
parameters of a parametric distribution $p(z_t|\theta_t)$.  For
example, $p(z_t|\theta_t)$ could be a Gaussian and $y(\cdot)$ output the
Gaussian's mean and variance as $\theta_t$.
Fig.~\ref{fig:archzoomedouttrain} illustrates the overall
architecture (with additions for $\ctofar$ noted).
Information about the conditioning range $z_1 \ldots z_T$ is conveyed
through the state of the network at time $T$ (i.e., $h_{T}$).
From an encoder-decoder
perspective~\cite{sutskever2014sequence,bahdanau2015neural}, DeepAR
uses the same network to encode and decode.

For training, each series is sliced into multiple \emph{windows},
i.e., conditioning+prediction ranges at different start points.
Windows are normalized using their conditioning ranges, and
parameters are fit to minimize NLL of prediction-range outputs.
To generate a forecast for a given (normalized) conditioning range,
DeepAR draws samples $\zhat_{T+1} \ldots \zhat_{T+N}$ in sequence from
$p(z_t|\theta_t)$ (Fig.~\ref{fig:archzoomedouttest}).
Sampled roll-outs are unnormalized, and by repeating this procedure
many times, a Monte Carlo estimate of~(\ref{eqn:forecast}) is
obtained, from which desired forecast quantiles can be derived (see, e.g.,
Fig.~\ref{fig:forecastexample}).

\subsection{Forecasting with $\ctofar$: $\ctofarRNN$}\label{subsec:ctofarrnn}

\begin{figure}
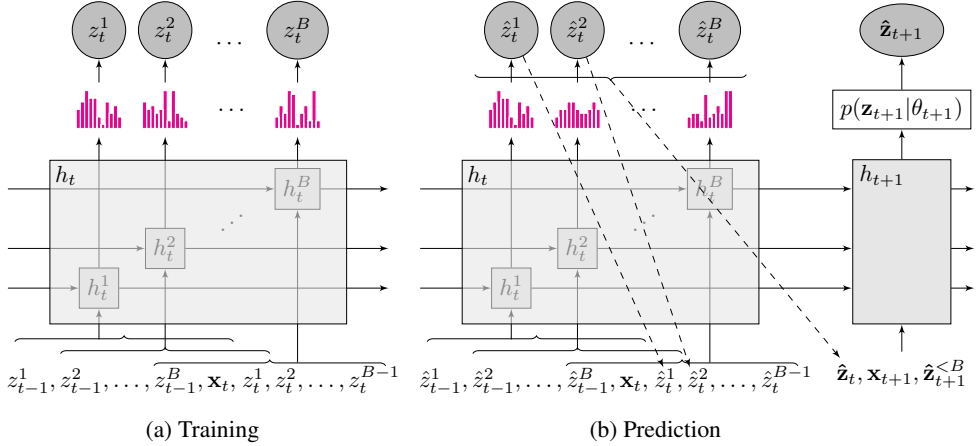

  \centering
      {\makebox[\textwidth][c]{
          \hspace{-13mm}
          \begin{subfigure}{\shrinkfigtwo\textwidth}
            \centering
            \scalebox{\tikzshrink}{
              {\input{tikz_figures/train_zoom_in.tex}}
            }
            \caption{Training\label{fig:archzoomedintrain}}
          \end{subfigure}
          \hspace{-10mm}
          \begin{subfigure}{\shrinkfigtwo\textwidth}
            \centering
            \scalebox{\tikzshrink}{
              {\input{tikz_figures/test_zoom_in.tex}}
            }
            \caption{Prediction\label{fig:archzoomedintest}}
          \end{subfigure}
      }}
      \caption{$\ctofarRNN$ model detailing high-level unit, $h_t$,
        from Fig.~\ref{fig:archzoomedout}. We train to minimize sum of
        NLL at each $z_t^i$ under a categorical with parameters given
        by a level-specific RNN, $h_t^i$.  We predict by sequentially
        sampling a discretization at each time step, $\zvec_t$, which
        is re-input at step $t+1$.\label{fig:archzoomedin}}
\end{figure}

We augment DeepAR by converting time series to their $\ctofar$
discretization $\zvec_t = \discrfunc(z_t)$.
Rather than generating $\theta_t$ in one shot at each time step, we
generate $\ctofar$ categoricals autoregressively \emph{within} each
time step (Fig.~\ref{fig:archzoomedin}).  Each generated bin index,
$z^{i-1}_t$, thus informs the distribution of the next bin index at
that time step $z^{i}_t$.
We also leverage information about discretized values at earlier time
steps
by replacing $\ctofar$'s classifier-based conditionals
(\S\ref{sec:c2far}, Fig.~\ref{fig:c2far}) with LSTMs, one for each
level in the $\ctofar$ hierarchy. Intuitively, when generating bin
index $z^i_t$, our inputs comprise both our current position in
higher-level bins (e.g. $z^{i-1}_t$) \emph{and} the $i$th-level index at
the previous time step ($z^i_{t-1}$), along with the previous LSTM
state, $h^i_{t-1}$.

We call the resulting system $\ctofarRNN$.  As with DeepAR,
$\ctofarRNN$ is trained to minimize NLL of observed outputs in
(normalized) prediction ranges (Fig.~\ref{fig:archzoomedintrain}).
During training (and when evaluating NLL of test sequences), \emph{at
  each time step}, all values are known and all distributions can be
computed \emph{in parallel}.  When predicting
(Fig.~\ref{fig:archzoomedintest}), we must sample each bin index
$z^i_t$ \emph{sequentially}.

Recall that $\ctofar$ has an implicit $B+1$ level, where a real value
is generated from a parametric distribution (Eqn.~(\ref{eqn:pareto})).
We do not use an RNN for this level; the uniform distributions are
fully-specified by the interval endpoints, while we generate Pareto
parameters ($\alpha_1$, $\alpha_2$) via a simple feed-forward neural
network (with a single hidden layer and softplus output
transformation).  To help inform the Pareto networks, we also encode
real-valued $z_{t-1}$ as an additional input/covariate.

\textbf{Complexity}.  Let $H$ be the number of RNN hidden units, $K$ a
constant number of bins per level, and $B$ the number of
levels ($I=K^B$ total intervals).  The complexity-per-timestep of
\emph{each RNN} is essentially the sum of the RNN's recurrence
operation, $H \times H$, and the projection of the recurrence to output bins, $K
\times H$.  For flat binnings, \emph{overall} complexity is typically
dominated by $K \times H$, while $B$-level $\ctofarRNN$ models are
dominated by $B \times H \times H$, i.e., the cost of running $B$ RNNs
in parallel.  Measurements of timing and memory consumption
(supplemental
Table~\ref{supp:tab:testing_time},~\ref{supp:tab:resources}) are well
explained by these observations and corresponding values for H, K,
and B (supplemental 
Table~\ref{supp:tab:tuning_results}).\footnote{From a discretization
  perspective, the optimal $K$ is 2, as rather than a softmax, we may
  use a logistic function thresholded at 0.5 to select high vs.\@ low
  bins, only computing $\log_2(I)$ total outputs.  However, complexity
  also depends on the neural architecture. As C2FAR-RNN operates $B$
  separate RNNs, it is more efficient to use a higher value of $K$
  (but not so high that computing output probabilities dominates) and
  a lower value of $B$.}

\textbf{Tuning}.  While $\ctofar$ is trained for NLL, it is
\emph{tuned} for a given target metric.
For forecasting, the metric is multi-step-ahead error.  Thus we
evaluate our forecasts during training by periodically running Monte
Carlo sampling on our validation set and computing multi-step-ahead
error.
For tasks such as anomaly detection, denoising, etc., we would train
for NLL and tune for an application-specific metric.
While there is no one-size-fits-all metric for evaluating generative
models, log-likelihood itself is sometimes regarded as the de facto
standard~\cite{theis2015note}.
Ironically, this is the one loss function $\ctofar$ can \emph{not}
tune for, at least not on discrete data, because of the density spike
issue (\S\ref{subsec:challenges}); if we tune directly for NLL, the
tuner chooses more and more bins and levels, leading to narrower and
higher spikes.
If we \emph{must} tune for log-likelihood, we could use discrete
distributions in the $B+1$ level.  We could also add uniform noise, as
other approaches do (but note in prior work this is required to enable
\emph{training}, not \emph{tuning}, on discrete data).  NLL
evaluations are explored in supplemental
\S\ref{supp:sec:likelihoodexp}.

\subsection{Limitations and broader impact of $\ctofar$ forecasting}\label{subsec:limitations}

Every B-level $\ctofar$ model has an equally
\emph{expressive}~\cite{raghu2017expressive} flat counterpart with a
single categorical over all the fine-grained intervals; given
unlimited training data, $\ctofar$ may therefore not offer modeling
benefits over flat binning.
Conversely, when there is limited data, simple parametric
distributions, with fewer parameters, may generalize better than
$\ctofar$.
Also, given it has extra hyperparameters (number of
levels/bins-per-level), and given tuning search space grows
exponentially with added hyperparameters, $\ctofar$ may not discover
optimum settings as quickly.
We investigate experimentally whether the benefits of $\ctofar$ outweigh
these drawbacks on real-world data (\S\ref{subsec:empirical}).

$\ctofar$ is more \emph{complex} than flat binning, but whether it is
less \emph{efficient} depends on implementation.
Compared to their 2-level coarse-to-fine model,
WaveRNN~\cite{kalchbrenner2018efficient} found flat binning required
``significantly more parameters, memory and compute.''  We find
$\ctofar$ to run slower, but with less memory
(Tables~\ref{supp:tab:testing_time} and~\ref{supp:tab:resources} in
the supplemental).
If flat binning was truly more efficient,
we could always generate a large volume of data from a $\ctofar$ model
and \emph{uptrain}~\cite{bucilua2006model,petrov2010uptraining} a flat
model on it.

\textbf{Broader impact.}  By enabling simultaneous modeling of
discrete and continuous series, without human involvement, $\ctofar$
is a step toward a universal forecast model.  Large $\ctofar$ models
could be trained on vast quantities of diverse series (similar to
efforts in text~\cite{devlin2018bert,liu2019roberta,yang2019xlnet} and
vision~\cite{simonyan2014very,szegedy2015going,he2016deep}).  Such
models could be fine-tuned for new domains, and help make
highly-accurate forecasting systems more widely-used, improving
decision making and resource allocation.

There are also risks to this approach.  Very large models have
environmental and financial costs~\cite{bender2021dangers}, which may
be unnecessary when smaller models suffice.
A universal model may be used more easily by those with less expertise
and this may lead to misuse; for example, \emph{automation bias} has
been shown to disproportionately affect those with less domain
expertise~\cite{bond2018automation}.
Recommended usage and possible misuse should be documented through
artifacts such as model cards~\cite{mitchell2019model} and
datasheets~\cite{gebru2021datasheets}.

\section{Experiments}\label{sec:experiments}

$\ctofar$ is implemented in \texttt{PyTorch}~\cite{paszke2019pytorch},
using a 2-layer LSTM~\cite{hochreiter1997long} with intra-layer
dropout~\cite{srivastava2014dropout,melis2017state}, trained via
Adam~\cite{kingma2014adam}.
Notation $\ctofarB$ refers to a $B$-level $\ctofar$ model.  We
evaluate:
\begin{itemize}[leftmargin=*]
\item $\ctofarone$: essentially the standard flat-binning
  approach~\cite{rabanser2020effectiveness}, but with Pareto
  tails~\cite{ehrlich2021spliced}
\item $\ctofartwo$: a two-level $\ctofarRNN$ model
\item $\ctofarthree$: a three-level $\ctofarRNN$ model
\end{itemize}

\subsection{Distribution recovery from synthetic data}\label{subsec:distribution}

\begin{figure}
  \centering
  {\makebox[\textwidth][c]{
      \begin{subfigure}{\shrinkfigtwo\textwidth}
        {\includegraphics[width=\textwidth]{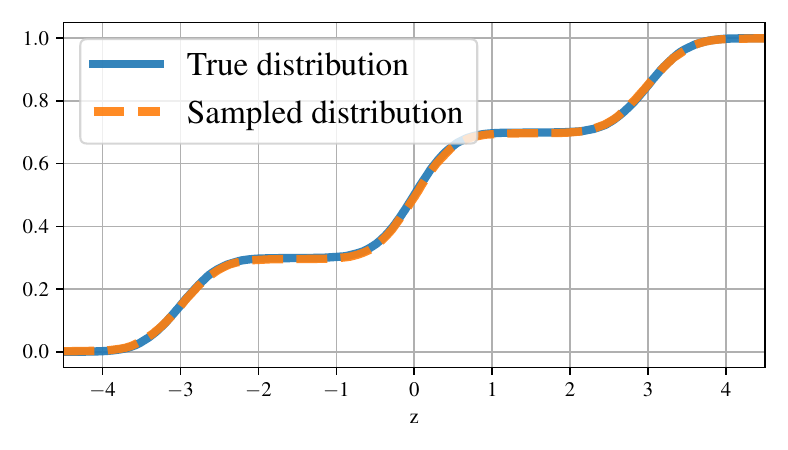}}
      \end{subfigure}
      \begin{subfigure}{\shrinkfigtwo\textwidth}
        {\includegraphics[width=\textwidth]{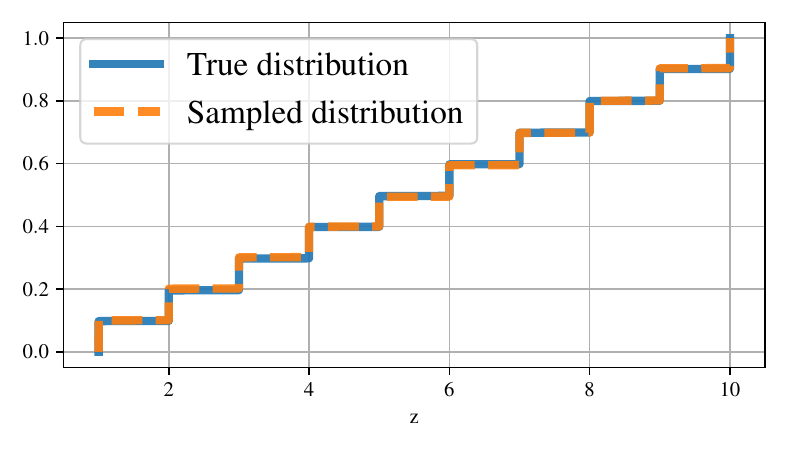}}
      \end{subfigure}
  }}
  \vspace{-2mm}
  \caption{Distribution recovery: CDFs for a Gaussian mixture (left)
    and discrete uniform over $1 \ldots 10$ (right). After training
    $\ctofarRNN$s on each, we sample paths and plot the \emph{sampled}
    CDF\@. $\ctofar$ recovers the true CDF much better than SQF-RNN or
    IQN-RNN (cf. Fig.~3 in~\cite{gasthaus2019probabilistic}, Fig.~2
    in~\cite{gouttes2021probabilistic}).\label{fig:recovery}}
\end{figure}

We first evaluate $\ctofarRNN$'s ability to recover the distribution
of synthetic data,
following~\cite{gasthaus2019probabilistic,gouttes2021probabilistic}.
We generate data exactly as in~\cite{gasthaus2019probabilistic},
creating 500 series with a year's worth of hourly observations, each
value sampled independently from a 3-component Gaussian mixture with
weights $[0.3, 0.4, 0.3]$, means $[-3, 0, 3]$ and standard deviation
of 0.4.
To illustrate $\ctofarRNN$'s ability to model discrete data, we create
an equivalent dataset where every element is drawn from a discrete
uniform, $\setU\{1,10\}$.

On each dataset, we trained a $\ctofarthree$ model with 20 bins per
level and hyperparameters based on $\elec$ experiments (below).

\textbf{Results.} In prior work,
SQF-RNN~\cite{gasthaus2019probabilistic} and
IQN-RNN~\cite{gouttes2021probabilistic} struggled to recover the
Gaussian mixture, but $\ctofar$ can fit both distributions perfectly
(Fig.~\ref{fig:recovery}).
$\ctofar$ is evidently the state-of-the-art in capturing complex,
multi-modal densities without any prior information.

\subsection{Empirical study on real-world data}\label{subsec:empirical}

\textbf{Datasets.}
We evaluate $\ctofar$ forecasting models on the following datasets:
\begin{itemize}[leftmargin=*]
\item $\elec$: hourly electricity usage of 321
  customers~\cite{dua2017uci}, using \emph{discretized} version
  from~\cite{lai2018modeling}.
\item $\traffic$: real-valued hourly occupancy from 0 to 1 for 862 car
  lanes~\cite{dua2017uci}, using version from~\cite{yu2016temporal}.
\item $\wiki$, daily integer number of hits for 9535 pages, first used
  in~\cite{gasthaus2019probabilistic}, using same version as
  in~\cite{gouttes2021probabilistic}.
\item $\azure$, (discrete) hourly usage of virtual machine (VM)
  flavors and flavor groupings (by tenant, deployment type, etc., in
  units of VCPUs or GB of memory) in a large public cloud, based on
  data from~\cite{cortez2017resource}. We release this dataset
  publicly as a paper supplement (see
  supplementary~\S\ref{supp:subsec:azure} for further details).
\end{itemize}

For $\azure$, we use 20 days as training, 3 days for validation, and 3
final days for testing.  Other splits are as
in~\cite{salinas2019high}.  We forecast 24 hours for hourly series, 30
days for $\wiki$, and use rolling evals as
in~\cite{gouttes2021probabilistic}.

\textbf{Metrics.}  To evaluate point forecasts, we output the median
of our forecast distribution and compute \emph{normalized deviation}
(ND) from true values.
We evaluate probabilistic forecasts using \emph{weighted quantile
  loss} (wQL),
evaluated at forecast quantiles $\{0.1, 0.2 \ldots 0.9\}$, as
in~\cite{rabanser2020effectiveness}.  For non-probabilistic baselines
$\naive$ and $\seasonalnaive$ (described below), note wQL reduces to
ND, similarly to how CRPS reduces to absolute
error~\cite[\S4.2]{gneiting2007strictly}.
We also test the \emph{calibration} and
\emph{sharpness}~\cite{wen2017multi} of each system, measuring
coverage of true values within particular percentiles (e.g. from 10\%
to 90\%, coverage closer to 80\% is better) and normalized width of
this interval (lower is better). For a target coverage of X\%, we
refer to these metrics together as \emph{CovX} (e.g. Cov80).  Scores
at extreme percentiles (e.g., Cov99\%) help evaluate modeling of
distribution tails.

\textbf{Baselines.} We compare our model to the following baselines:
\begin{itemize}[leftmargin=*]
\item $\naive$~\cite{hyndman2018forecasting}: outputs the
  last-observed historical value at all forecast horizons
\item $\seasonalnaive$~\cite{hyndman2018forecasting}: at each horizon,
  outputs the value at the most-recently-observed \emph{season}
  matching the season at that horizon.  We thus output the value at
  the most-recently-observed matching hour-of-day for hourly series,
  and matching day-of-week for the daily dataset ($\wiki$).
\item $\ets$~\cite{hyndman2008forecasting}: probabilistic state space
  models based on exponential smoothing, as implemented in
  \texttt{statsmodels}~\cite{seabold2010statsmodels}.  For each
  dataset, we tune (on validation data) whether to include
  \emph{seasonality} and \emph{trend} (\emph{damped} or
  \emph{undamped}) terms, and whether to use \emph{estimated} or
  \emph{heuristic} initialization.
\item $\deepargaussian$: our implementation of DeepAR with a Gaussian
  output distribution; Gaussian outputs are common in practice, even
  on discrete
  data~\cite{li2019enhancing,salinas2020deepar,chen2020probabilistic}
\end{itemize}

\textbf{Tuning.}  Deeper $\ctofar$ models include shallower models as
special cases.
We therefore address the question: does $\ctofar$'s improved modeling
outweigh the wider tuning required?
For fair comparison, we establish a \emph{parameter
  budget}~\cite{melis2017state} of 1M parameters; after sampling other
hyperparameters, we restrict the possible number of LSTM hidden units
so that this cap is enforced.
We also restrict the number of tuning trials (on validation data) to
100 for each system.\footnote{In the prior work that mentions
  hyperparameter tuning, it is common to compare systems tuned with a
  fixed number of tuning trials, whether as part of grid
  search~\cite{zhou2021informer,zhu2021mixseq,rangapuram2021end,salinas2019high}
  or other procedure (e.g., hyperopt in~\cite{challu2022n}, random
  search in~\cite{lim2021temporal}). Training times are on the same
  order for all our trained systems (supplemental
  Table~\ref{supp:tab:training_time}).}
We tune directly for normalized deviation via
Optuna~\cite{akiba2019optuna}, with
TPESampler~\cite{bergstra2011algorithms}, using MedianPruning and early
stopping.

{\begin{table}
  \caption{ND, wQL, Cov80, and Cov99 for our implementations (top),
    results from~\cite{rabanser2020effectiveness} (middle,
    denoted~$\dagger$) and~\cite{gouttes2021probabilistic} (bottom,
    denoted~$\ddagger$), where available.  In all cases, flat binned
    $\ctofarone$ improves on $\deepargaussian$, while deeper
    $\ctofartwo$ likewise improves over $\ctofarone$.  Results are
    generally superior to prior state-of-the-art output distributions
    in RNN-based forecasting.\label{tab:results}} \centering
  \footnotesize
  \begin{tabular}{@{}lcccccccccc@{}}
    \toprule
    \multicolumn{1}{c}{}  & \multicolumn{4}{|c|}{ND\%} & \multicolumn{4}{|c}{wQL\%} & \multicolumn{1}{|c}{Cov80\%} & \multicolumn{1}{|c}{Cov99\%}                         \\
    \multicolumn{1}{c}{}  & \multicolumn{1}{|c}{$\elec$}  & $\traffic$    & $\wiki$       & $\azure$            & \multicolumn{1}{|c}{$\elec$}  & $\traffic$    & $\wiki$      & $\azure$ & \multicolumn{1}{|c}{$\azure$} & \multicolumn{1}{|c}{$\azure$}     \\
    \midrule
    $\naive$                 & 40.8          & 73.6          & 35.7          & 3.49          & 40.8 & 73.6          & 35.7          & 3.49 & - & -          \\
    $\seasonalnaive$        & 6.97          & 25.1          & 33.2          & 3.67          & 6.97 & 25.1          & 33.2          & 3.67 & - & -          \\
    $\ets$                   & 8.61          & 33.3          & 34.3          & 3.46          & 8.40 & 31.5          & 32.5          & 2.97 & 85.5/10.5 & 96.3/\textbf{20.3}          \\
    $\deepargaussian$     & 7.05          & 16.1          & 43.8          & 3.60          & 5.60 & 13.7          & 54.7          & 3.06 & 89.9/16.9 &    98.0/37.7         \\
    $\ctofarone$          & 6.14          & 13.0          & 24.6          & 2.95          & 4.87 & 10.7          & 21.3          & 2.41 & 83.6/\textbf{8.3} &    98.5/32.2          \\
    $\ctofartwo$          & 6.09          & \textbf{12.9} & 24.2          & 2.86          & 4.83 & \textbf{10.6} & \textbf{21.0} & 2.31 & \textbf{79.0}/8.5 &    98.4/29.1          \\
    $\ctofarthree$        & \textbf{6.00} & 13.3          & \textbf{24.1} & \textbf{2.77} & \textbf{4.76} & 10.9          & \textbf{21.0} & \textbf{2.27} &    86.0/8.9 &    \textbf{98.6}/32.7 \\ \midrule
    DeepAR-Binned$\dagger$         & 8.21          & 23.2          & 94.6          & -             & 6.47 & 18.8          & 84.7          & -  & -  & -          \\
    DeepAR-StudentT$\dagger$       & 6.95          & 14.6          & 26.9          & -             & 5.71 & 12.2          & 23.8          & -    & - & -        \\ \midrule
    IQN-RNN$\ddagger$               & 7.40          & 16.8          & \textbf{24.1} & -             & -    & -             & -             & -   & - & -     \\
    SQF-RNN$\ddagger$               & 9.70          & 18.6          & 32.8          & -             & -    & -             & -             & -   & - & -          \\
    DeepAR-StudentT$\ddagger$       & 7.80          & 21.6          & 27.0          & -             & -    & -             & -             & -       & - & -      \\ \bottomrule
  \end{tabular}
\end{table}
}

\textbf{Results}.  Table~\ref{tab:results} confirms that deeper
$\ctofar$ models improve over flat binnings.
In only one case ($\traffic$) did a $\ctofar$ model not improve over
flat binning.  In this case, $\ctofarthree$, with a larger tuning
search space, also failed to find a superior setting of
hyperparameters on validation data; going forward, using the same
number of bins at each level (i.e., a constant $K$) could enable
deeper $\ctofar$ models without additional tuning.
Contrary to prior work, we also find flat binning much more effective
than standard parametric distributions.
We attribute this difference to our use of systematic tuning; prior
work used a fixed 1024 bins~\cite{rabanser2020effectiveness}, while
our tuner often selected quite fewer bins for $\ctofarone$
(supplemental Table~\ref{supp:tab:tuning_results}).
In terms of both standard and extreme percentiles (Cov80\% and Cov99\%
in Table~\ref{tab:results}), we find $\ctofar$ models are both better
calibrated than $\deepargaussian$ (being twice as close to the desired
coverage) while also having a sharper prediction interval width.

$\seasonalnaive$ is surprisingly competitive with $\deepargaussian$
(except on $\traffic$, which has both daily and weekly seasonality).
This has been observed previously (e.g., Table~1
in~\cite{alexandrov2020gluonts}); indeed, one of the authors of DeepAR
and GluonTS has remarked that ``a seasonal naive model performs almost
as well as DeepAR or other deep models on [$\elec$ and $\traffic$],
and without additional covariates, I suspect it's almost impossible to
perform significantly better in terms of point
predictions''~\cite{gasthaus2020github}.  But our results show that
C2FAR \emph{can} significantly improve over DeepAR and
$\seasonalnaive$ without using additional covariates, but rather with
a better output model.

\begin{figure}
  \vspace{-1mm}
  \centering {\makebox[\textwidth][c]{
      \hspace{-3mm}
      \begin{subfigure}{\shrinkfigthree\textwidth}
        \begin{tikzpicture}
          \node (img1) {\includegraphics[width=\textwidth]{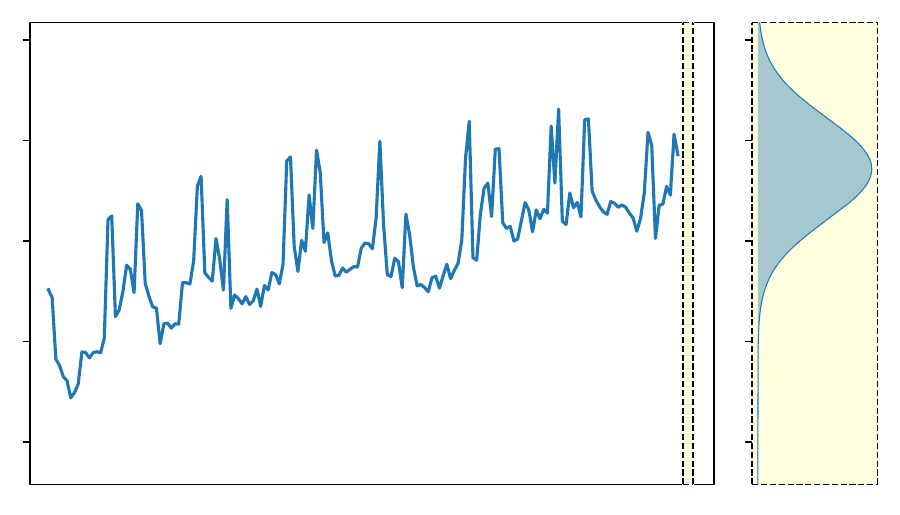}};
          \node[below of=img1, align=center, yshift=0.94cm, xshift=-2.12cm, node distance=0cm, anchor=west] {\scriptsize $\deepargaussian$ ($\azure$)};
        \end{tikzpicture}
      \end{subfigure}
      \hspace{-1mm}
      \begin{subfigure}{\shrinkfigthree\textwidth}
        \begin{tikzpicture}
          \node (img1) {\includegraphics[width=\textwidth]{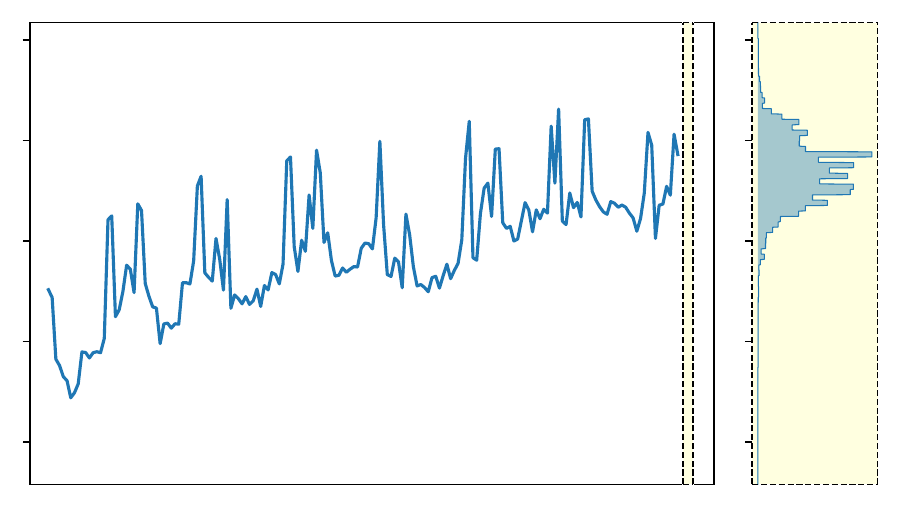}};
          \node[below of=img1, align=center, yshift=0.94cm, xshift=-2.12cm, node distance=0cm, anchor=west] {\scriptsize $\ctofarone$ ($\azure$)};
        \end{tikzpicture}
      \end{subfigure}
      \hspace{-1mm}
      \begin{subfigure}{\shrinkfigthree\textwidth}
        \begin{tikzpicture}
          \node (img1) {\includegraphics[width=\textwidth]{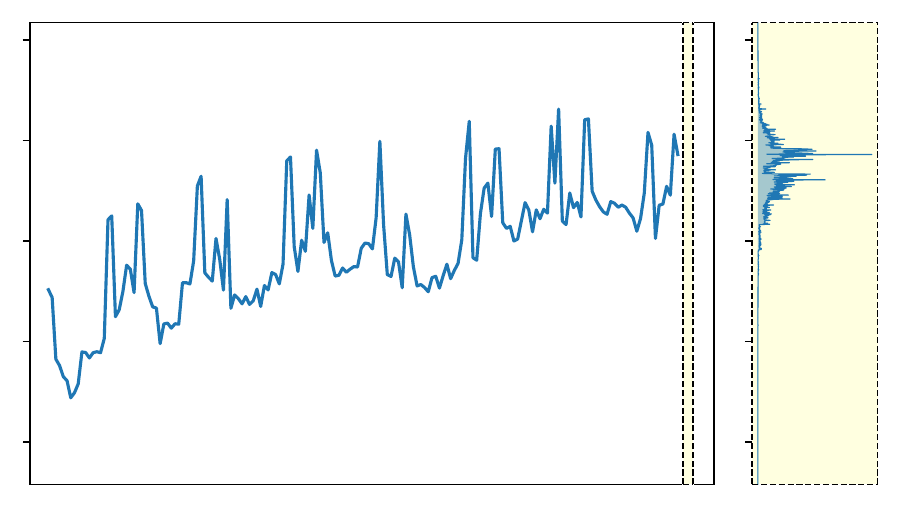}};
          \node[below of=img1, align=center, yshift=0.94cm, xshift=-2.12cm, node distance=0cm, anchor=west] {\scriptsize $\ctofartwo$ ($\azure$)};
        \end{tikzpicture}
      \end{subfigure}
  }} \\
  \vspace{-3mm}
  {\makebox[\textwidth][c]{
      \hspace{-3mm}
      \begin{subfigure}{\shrinkfigthree\textwidth}
        \begin{tikzpicture}
          \node (img1) {\includegraphics[width=\textwidth]{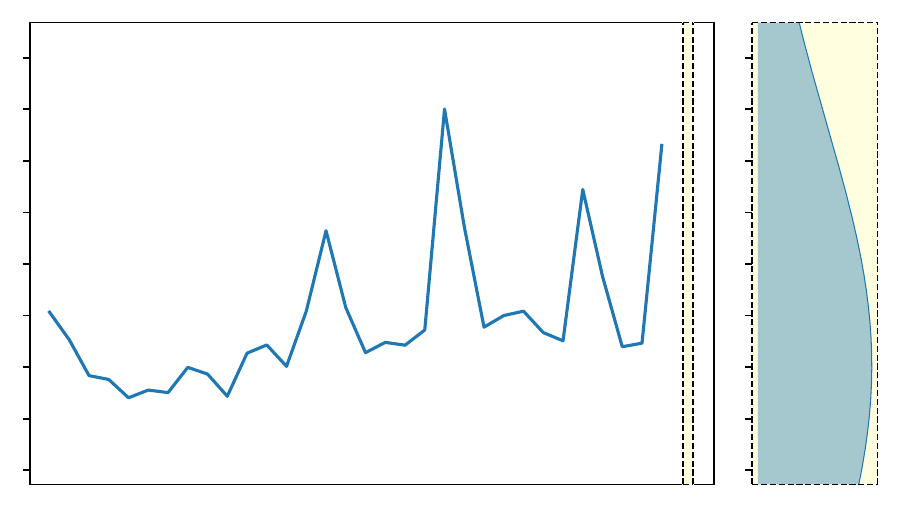}};
          \node[below of=img1, align=center, yshift=0.96cm, xshift=-2.12cm, node distance=0cm, anchor=west] {\scriptsize $\deepargaussian$ ($\wiki$)};
        \end{tikzpicture}
      \end{subfigure}
      \hspace{-1mm}
      \begin{subfigure}{\shrinkfigthree\textwidth}
        \begin{tikzpicture}
          \node (img1) {\includegraphics[width=\textwidth]{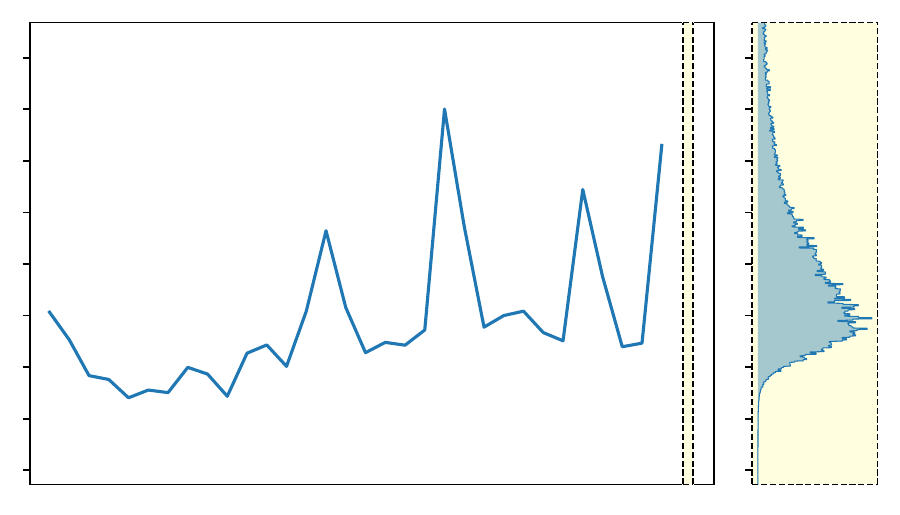}};
          \node[below of=img1, align=center, yshift=0.96cm, xshift=-2.12cm, node distance=0cm, anchor=west] {\scriptsize $\ctofarone$ ($\wiki$)};
        \end{tikzpicture}
      \end{subfigure}
      \hspace{-1mm}
      \begin{subfigure}{\shrinkfigthree\textwidth}
        \begin{tikzpicture}
          \node (img1) {\includegraphics[width=\textwidth]{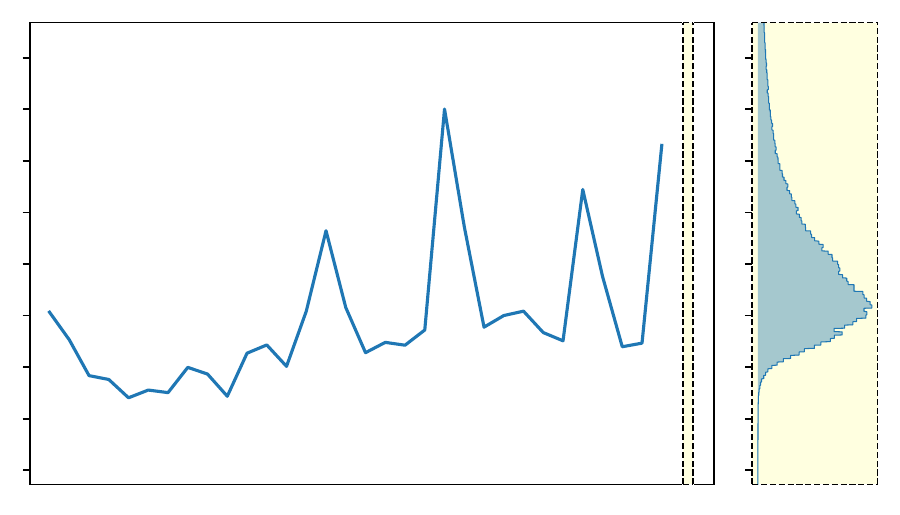}};
          \node[below of=img1, align=center, yshift=0.96cm, xshift=-2.12cm, node distance=0cm, anchor=west] {\scriptsize $\ctofartwo$ ($\wiki$)};
        \end{tikzpicture}
      \end{subfigure}
  }}
  \vspace{-3mm}
  \caption{Distributions over possible next values.  To cover likely
    values, Gaussians (left) also cover many unlikely ones.  Flat
    binnings (middle) are optimized to either use fewer bins, but
    suffer in precision (top), or many bins but suffer in noise
    (bottom).  $\ctofartwo$ (right) is able to place high probability
    on a repeat of the previous value (top) or generate smooth
    distributions (bottom).\label{fig:outputs}}
\end{figure}
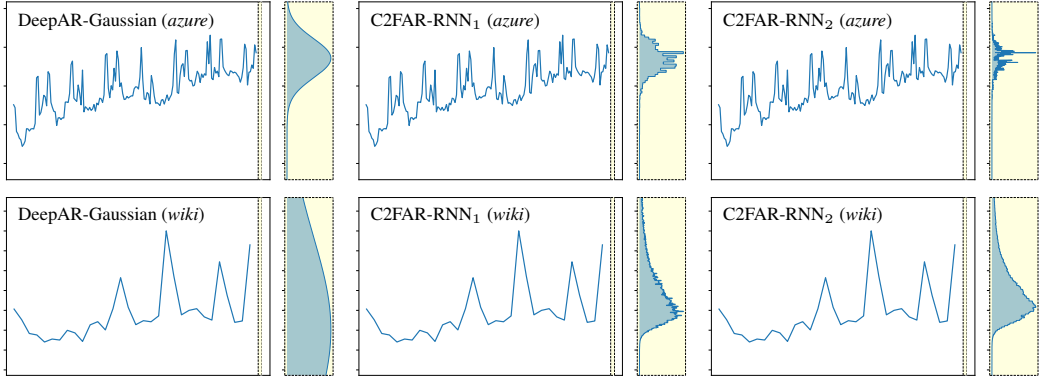

Analyzing the distributions, we find $\ctofar$ models are able to
achieve good \emph{precision} (having many intervals), while
simultaneously learning smooth distributions (Fig.~\ref{fig:outputs}).
Flat binnings must first \emph{learn} ``that a value of 128 is close
to a value of 127 or 129''~\cite{salimans2017pixelcnn++}, while for
$\ctofar$ models, intervals will implicitly be close in probability
because they are in the same coarser bins.

\begin{figure}
  \vspace{-1mm}
  \centering {\makebox[\textwidth][c]{
      \hspace{-3mm}
      \begin{subfigure}{\shrinkfigtwo\textwidth}
        \begin{tikzpicture}
          \node (img1) {\includegraphics[width=\textwidth]{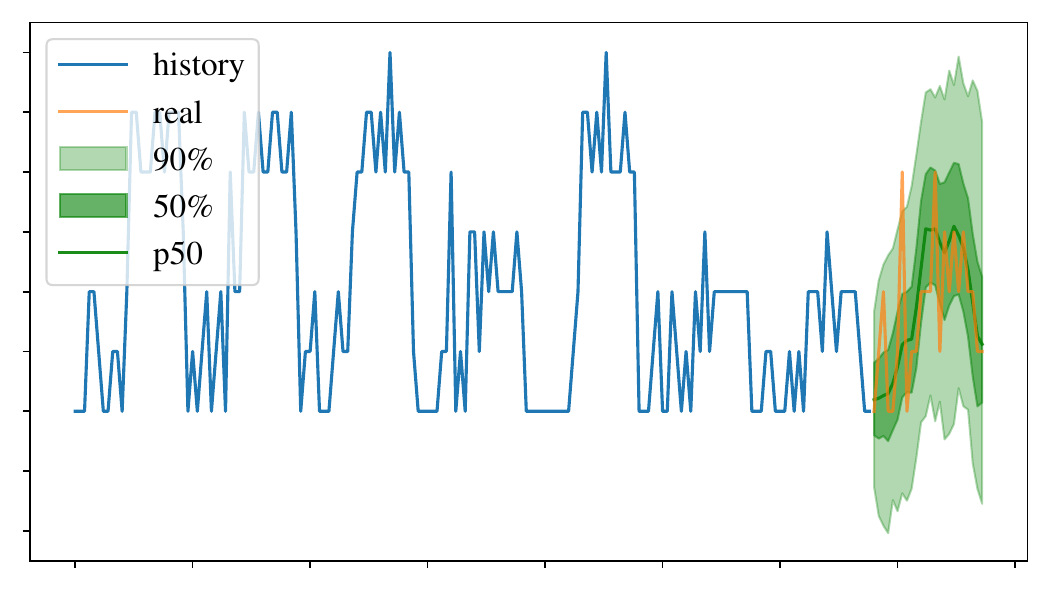}};
          \node[below of=img1, align=center, yshift=-1.46cm, xshift=-3.12cm, node distance=0cm, anchor=west] {\scriptsize $\deepargaussian$ ($\elec$)};
        \end{tikzpicture}
      \end{subfigure}
      \hspace{-1mm}
      \begin{subfigure}{\shrinkfigtwo\textwidth}
        \begin{tikzpicture}
          \node (img1) {\includegraphics[width=\textwidth]{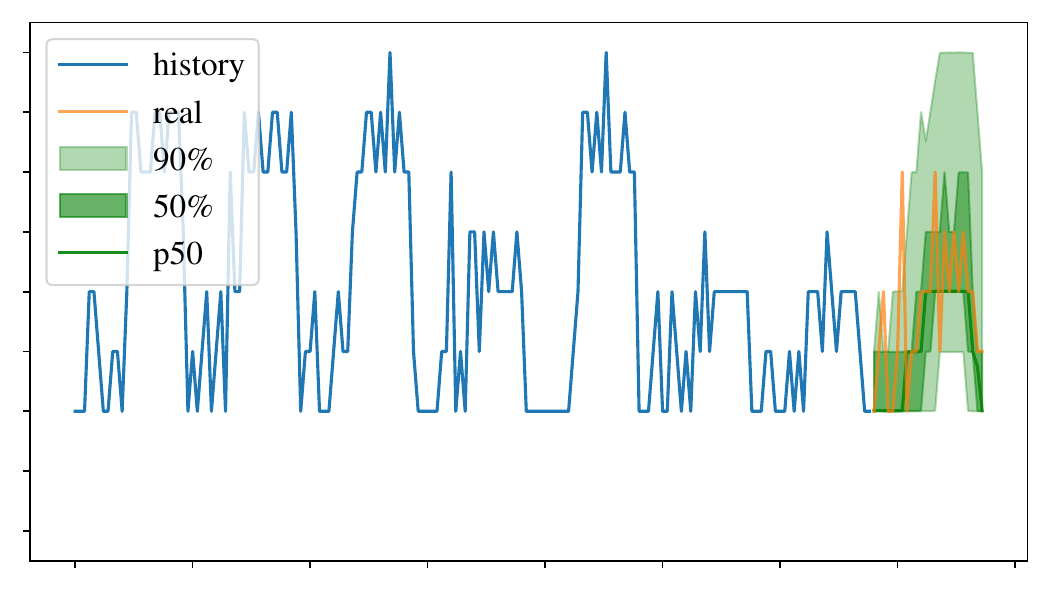}};
          \node[below of=img1, align=center, yshift=-1.46cm, xshift=-3.12cm, node distance=0cm, anchor=west] {\scriptsize $\ctofartwo$ ($\elec$)};
        \end{tikzpicture}
      \end{subfigure}
  }}
  \vspace{-3mm}
  \caption{Forecast for an $\elec$ time series.  $\deepargaussian$
    (left) yields a reasonable p50, but its lower percentiles span too
    low.  $\ctofartwo$ yields better lower quantiles.  $\ctofar$
    percentiles also suggest high-fidelity roll-outs, i.e., samples
    that closely mimic the discretized dynamics of the
    series.\label{fig:forecastexample}}
\end{figure}

Autoregressive models have a known disconnect between training, when
true values are used as autoregressive inputs, and generation, when
samples are used~\cite{bengio2015scheduled}.
Our results suggest that such error accumulation may be especially
problematic when a model is trained on discrete or mixed data, but
uses standard continuous parametric outputs (as in $\deepargaussian$).
Precise $\ctofar$ models not only provide better forecast quantiles,
they enable higher-fidelity samples to be recursively fed back in as
autoregressive inputs (Fig.~\ref{fig:forecastexample}).
Analyzing error by forecast horizon (supplemental
\S\ref{supp:subsec:horizons}), we find that at shorter horizons,
$\deepargaussian$ often performs similarly to $\ctofar$, but at later
horizons, the gap widens.
So while other solutions to error accumulation
exist~\cite{taieb2015bias,wen2017multi,wu2020adversarial}, one
effective approach is evidently to generate and recurse on
higher-fidelity outputs using $\ctofar$.

\section{Conclusion}

We presented $\ctofar$, a new method for density modeling.  $\ctofar$
reduces the generative process to a sequence of classifications over a
hierarchical, discretized representation, with special handling of
data outside the binning range.
$\ctofar$ can be applied to a variety of neural architectures, including
RNN-based probabilistic forecasting, where it achieves
state-of-the-art results when recovering synthetic distributions and
forecasting real-world data.
We show binned models (whether flat or, especially, coarse-to-fine)
are superior to standard distributions --- if binning precision is
tuned.

Analysis shows that $\ctofar$ can successfully model complex,
multi-modal densities, in real or discrete data, without any prior
information.
This flexibility enables improved modeling for a variety of time
series use cases, including forecasting, anomaly detection,
interpolation, compression, denoising, and generating high-fidelity
samples.  It also enables development of large-scale forecasting
models trained on diverse datasets, i.e., it is a step toward a
universal neural forecaster.

\bibliographystyle{ACM-Reference-Format}
\bibliography{futurenet}

\section*{Checklist}

\begin{enumerate}

\item For all authors...
\begin{enumerate}
  \item Do the main claims made in the abstract and introduction accurately reflect the paper's contributions and scope?
    \answerYes{}
  \item Did you describe the limitations of your work?
    \answerYes{In particular, \S\ref{subsec:limitations}}
  \item Did you discuss any potential negative societal impacts of your work?
    \answerYes{\S\ref{subsec:limitations}}
  \item Have you read the ethics review guidelines and ensured that your paper conforms to them?
    \answerYes{}
\end{enumerate}

\item If you are including theoretical results...
\begin{enumerate}
  \item Did you state the full set of assumptions of all theoretical results?
    \answerNA{}
        \item Did you include complete proofs of all theoretical results?
    \answerNA{}
\end{enumerate}

\item If you ran experiments...
\begin{enumerate}
  \item Did you include the code, data, and instructions needed to reproduce the main experimental results (either in the supplemental material or as a URL)?
    \answerNo{The full codebase is proprietary, however the core code for C2FAR
      is shared at \url{https://github.com/huaweicloud/c2far_forecasting}.  All datasets are either shared as a paper contribution, or already public.}
  \item Did you specify all the training details (e.g., data splits, hyperparameters, how they were chosen)?
    \answerYes{See \S\ref{subsec:empirical} and paper supplement.}
        \item Did you report error bars (e.g., with respect to the random seed after running experiments multiple times)?
    \answerYes{See paper supplement.}
        \item Did you include the total amount of compute and the type of resources used (e.g., type of GPUs, internal cluster, or cloud provider)?
    \answerYes{See paper supplement.}
\end{enumerate}

\item If you are using existing assets (e.g., code, data, models) or curating/releasing new assets...
\begin{enumerate}
  \item If your work uses existing assets, did you cite the creators?
    \answerYes{\S\ref{subsec:empirical}}
  \item Did you mention the license of the assets?
    \answerYes{For shared data, see supplement}
  \item Did you include any new assets either in the supplemental material or as a URL\@?
    \answerYes{See paper supplement.}
  \item Did you discuss whether and how consent was obtained from people whose data you're using/curating?
    \answerYes{See paper supplement.}
  \item Did you discuss whether the data you are using/curating contains personally identifiable information or offensive content?
    \answerYes{See paper supplement.}
\end{enumerate}

\item If you used crowdsourcing or conducted research with human subjects...
\begin{enumerate}
  \item Did you include the full text of instructions given to participants and screenshots, if applicable?
    \answerNA{}
  \item Did you describe any potential participant risks, with links to Institutional Review Board (IRB) approvals, if applicable?
    \answerNA{}
  \item Did you include the estimated hourly wage paid to participants and the total amount spent on participant compensation?
    \answerNA{}
\end{enumerate}

\end{enumerate}

\appendix

\section{Experimental details}

\subsection{Architecture}

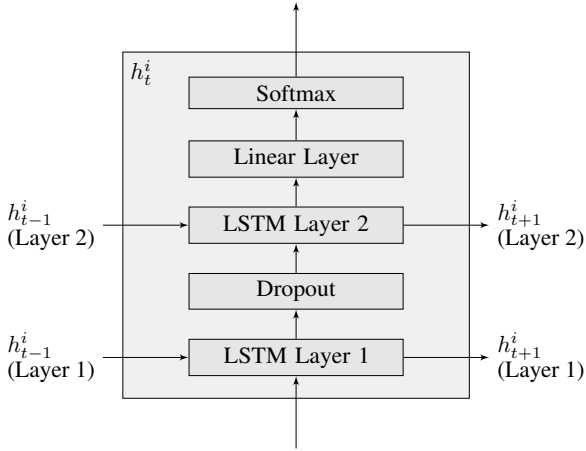
\begin{figure}[h]
  \centering
  \scalebox{\tikzshrink}{
    {\begin{tikzpicture}
    \node [main2] (halo) {};
    \node [below right] at (halo.north west) {$h^i_{t}$};
    \node [gblock2, yshift=2cm] (Softmax) {Softmax};
    \node [gblock2, below of=Softmax] (LinearLayer) {Linear Layer};
    \node [gblock2, below of=LinearLayer] (LSTM2) {LSTM Layer 2};
    \node [halo, left of=LSTM2, xshift=-2.7cm, align=left] (h2t-1)
          {$h^{i}_{t-1}$ \\ (Layer 2)};
    \node [halo, right of=LSTM2, xshift=2.7cm, align=left] (h2t)
          {$h^{i}_{t+1}$ \\ (Layer 2)};
    \node [gblock2, below of=LSTM2] (Dropout) {Dropout};
    \node [gblock2, below of=Dropout] (LSTM1) {LSTM Layer 1};
    \node [halo, left of=LSTM1, xshift=-2.7cm, align=left] (h1t-1)
          {$h^{i}_{t-1}$ \\ (Layer 1)};
    \node [halo, right of=LSTM1, xshift=2.7cm, align=left] (h1t)
          {$h^{i}_{t+1}$ \\ (Layer 1)};
    \node [halo, below of=LSTM1, yshift=-.5cm] (input) {};
    \node [halo, above of=Softmax, yshift=.5cm] (output) {};
    \draw [draw, -latex'] (input) -- (LSTM1);
    \draw [draw, -latex'] (h1t-1) -- (LSTM1);
    \draw [draw, -latex'] (LSTM1) -- (h1t);
    \draw [draw, -latex'] (LSTM1) -- (Dropout);
    \draw [draw, -latex'] (Dropout) -- (LSTM2);
    \draw [draw, -latex'] (h2t-1) -- (LSTM2);
    \draw [draw, -latex'] (LSTM2) -- (h2t);
    \draw [draw, -latex'] (LSTM2) -- (LinearLayer);
    \draw [draw, -latex'] (LinearLayer) -- (Softmax);
    \draw [draw, -latex'] (Softmax) -- (output);
\end{tikzpicture}}
  }
  \caption{LSTM architecture for $h^i_{t}$: one level of $\ctofar$ at
    time $t$ and level $i$ (see Fig.~4 in main paper).  Network layers
    do include \emph{bias weights}.  The same number of hidden units
    are used in each LSTM layer and is given by hyperparameter
    \emph{n\_hidden}. Dropout probability is given by hyperparameter
    \emph{lstm\_dropout}.\label{fig:lstm_unit}}
\end{figure}

Regarding the sequence model, all $\ctofarRNN$ models use 2-layer
LSTMs~\cite{hochreiter1997long} with intra-layer
dropout~\cite{srivastava2014dropout,melis2017state}, as depicted in
Fig.~\ref{fig:lstm_unit}.  Multi-level $\ctofarRNN$ models use the
same LSTM architecture at each level, with the same number of hidden
units in each LSTM layer of each level.
We follow DeepAR~\cite{salinas2020deepar} in using the same network to
encode (i.e., process the conditioning range) and decode (i.e.,
generate values in the prediction range).  However, unlike DeepAR,
during training we only compute loss over the prediction range.

For generating the parameters of the Pareto distribution, we use a
feed-forward neural network with a single hidden layer (followed by a
\emph{softplus} output transformation).  The number of units in this
hidden layer is also controlled by the \emph{n\_hidden}
hyperparameter.

\subsection{Features and input/output preparation}

Toward our goal of universal forecasting (\S4.3 of the main paper), we
exclude covariates for series-specific meta data, series ``age'', and
lagged historical values (all of which are used
in~\cite{salinas2020deepar}).  We use min-max
scaling~\cite{rabanser2020effectiveness} to normalize conditioning
ranges prior to forecasting.
Autoregressive inputs are represented with 1-hot-encodings.

\subsection{Training and testing details}

We use Optuna~\cite{akiba2019optuna} for tuning, with the
TPESampler~\cite{bergstra2011algorithms}, and use both MedianPruning
(pruning unpromising trials compared to previous trials) and early
stopping (stopping trials when results no longer improve).
We stop early if we see \emph{n\_stop\_evals\_no\_improve} evaluations
without a new top score (currently set to 37, see
Table~\ref{tab:fixed}).

We filter instances with constant \emph{conditioning} ranges from
training and testing.  We do not oversample training instances from
the higher-amplitude series, as DeepAR does~\cite{salinas2020deepar}.

We vectorize across multiple series during both the computing of
likelihood (in training) and during the sampling of future values (in
prediction).  The number of series that we parallelize over is
referred to as our ``batch size'' (e.g. \emph{n\_train\_batch\_size}).

We also vectorize across different multi-step-ahead
rollouts during the Monte Carlo procedure to generate the forecast
distribution.
We use 500 separate rollouts during the forecasting process
(\emph{n\_rollouts} = 500), unless otherwise stated.
We compute rolling evaluations with a stride of 1, i.e., we forecast
and evaluate over overlapping prediction ranges, as
in~\cite{gouttes2021probabilistic}.

\subsection{Computational resources}

$\ctofar$ is implemented in \texttt{PyTorch}~\cite{paszke2019pytorch},
version \texttt{1.9.1+cu102}.
We use GPUs from Nvidia: four Tesla P100 GPUs with 16280MiB and two
Tesla K80 GPUs with 11441MiB.  To maximize the utilization of the
GPUs, we usually ran two trials in parallel on each of the six GPUs,
for 12 trials running in parallel in total at any one time for a given
tuning study.

\section{Distribution recovery}

Here we expand on the results in \S5.1 of the paper, evaluating each
of our our implemented systems on the task of recovering synthetic
distributions.

\subsection{Training details}

Rather than tuning the models for this (simple) task, we use a fixed
learning rate of 2e-2, a weight decay of 1e-6, and 64 hidden units,
which are settings that worked well during development experiments on
the $\elec$ validation set.  We also used an \emph{lstm\_dropout} of
1e-3 and training batch size of 1024 (as in Table~\ref{tab:fixed}).
The (normalized) binning extent is taken from -0.01 to 1.01.  We use a
conditioning range of 96 and a prediction range of 24,
following~\cite{gasthaus2019probabilistic}.  We take the final four
days as the testing period.

\subsection{Computational performance}

The time to execute the training runs varied between 53 minutes and 84
minutes for all systems on the synthetic Gaussian Mixture Model (GMM)
data, and between 10 minutes and 30 minutes for all systems on the
synthetic discrete data.

\subsection{Further results}

\begin{figure}
  \centering
  {\makebox[\textwidth][c]{
      \begin{subfigure}{\shrinkfigtwo\textwidth}
        {\includegraphics[width=\textwidth]{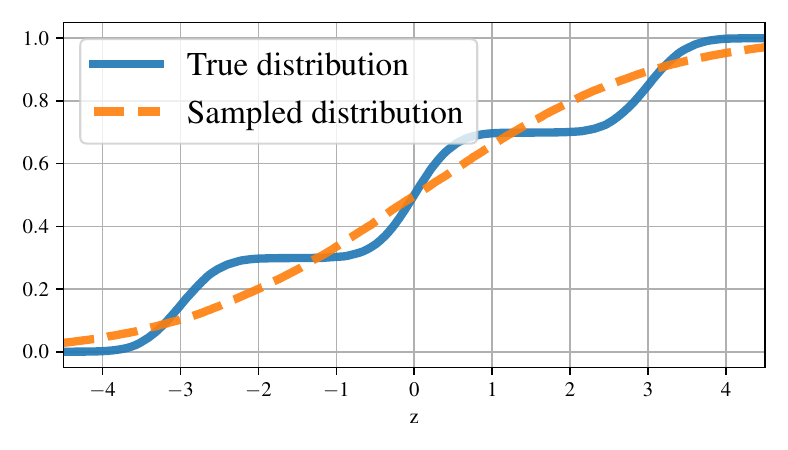}}
        \vspace{-5mm}
        \caption{GMM, $\deepargaussian$}
      \end{subfigure}
      \begin{subfigure}{\shrinkfigtwo\textwidth}
        {\includegraphics[width=\textwidth]{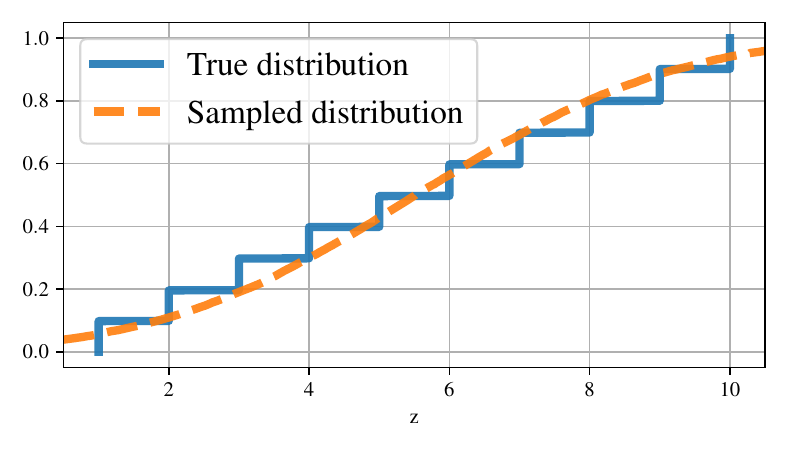}}
        \vspace{-5mm}
        \caption{Discrete, $\deepargaussian$}
      \end{subfigure}
  }}
  \\
  \vspace{+5mm}
  {\makebox[\textwidth][c]{
      \begin{subfigure}{\shrinkfigtwo\textwidth}
        {\includegraphics[width=\textwidth]{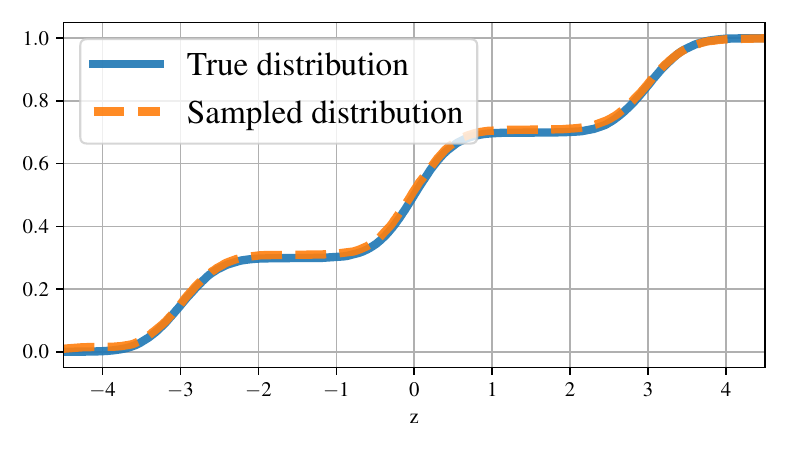}}
        \vspace{-5mm}
        \caption{GMM, $\ctofarone$ with $60$ bins}
      \end{subfigure}
      \begin{subfigure}{\shrinkfigtwo\textwidth}
        {\includegraphics[width=\textwidth]{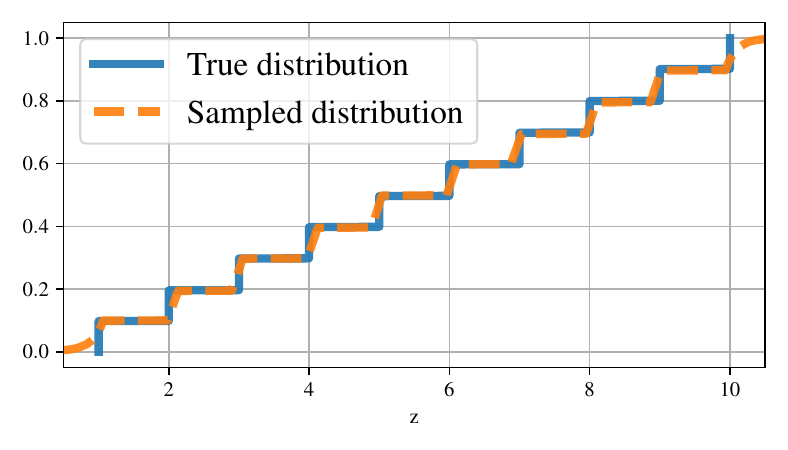}}
        \vspace{-5mm}
        \caption{Discrete, $\ctofarone$ with $60$ bins}
      \end{subfigure}
  }}
  \\
  \vspace{+5mm}
  {\makebox[\textwidth][c]{
      \begin{subfigure}{\shrinkfigtwo\textwidth}
        {\includegraphics[width=\textwidth]{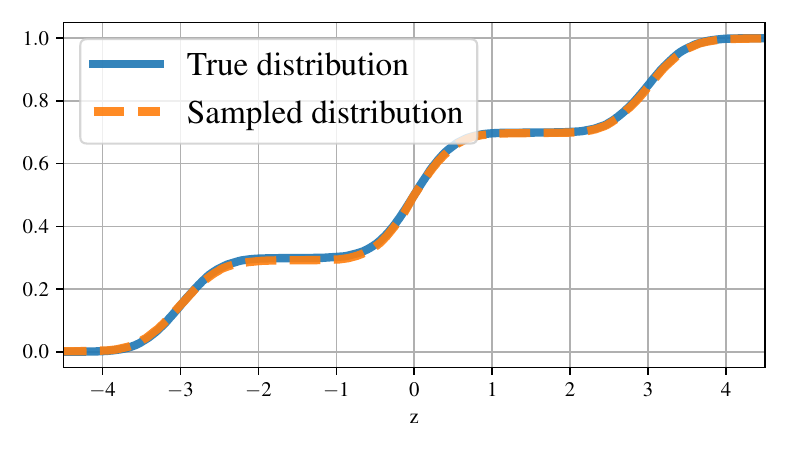}}
        \vspace{-5mm}
        \caption{GMM, $\ctofartwo$ with $30+30$ bins}
      \end{subfigure}
      \begin{subfigure}{\shrinkfigtwo\textwidth}
        {\includegraphics[width=\textwidth]{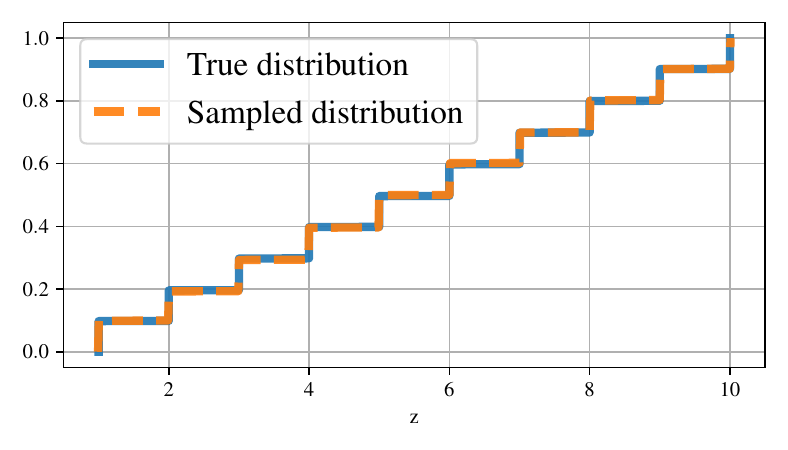}}
        \vspace{-5mm}
        \caption{Discrete, $\ctofartwo$ with $30+30$ bins}
      \end{subfigure}
  }}
  \\
  \vspace{+5mm}
  {\makebox[\textwidth][c]{
      \begin{subfigure}{\shrinkfigtwo\textwidth}
        {\includegraphics[width=\textwidth]{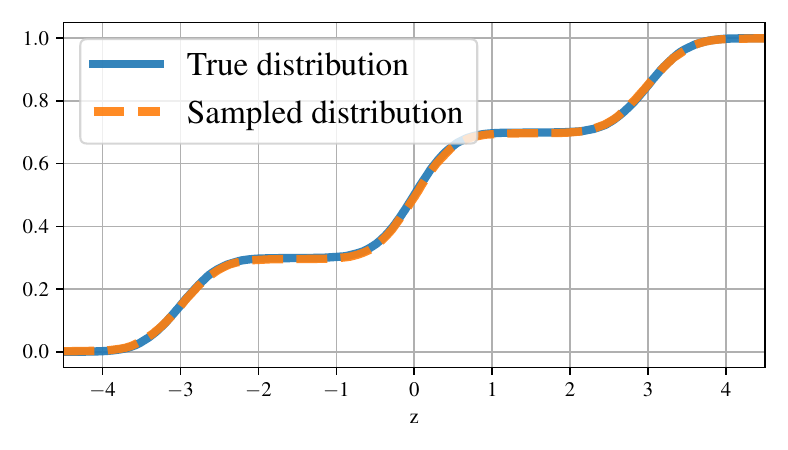}}
        \vspace{-5mm}
        \caption{GMM, $\ctofarthree$ with $20+20+20$ bins}
      \end{subfigure}
      \begin{subfigure}{\shrinkfigtwo\textwidth}
        {\includegraphics[width=\textwidth]{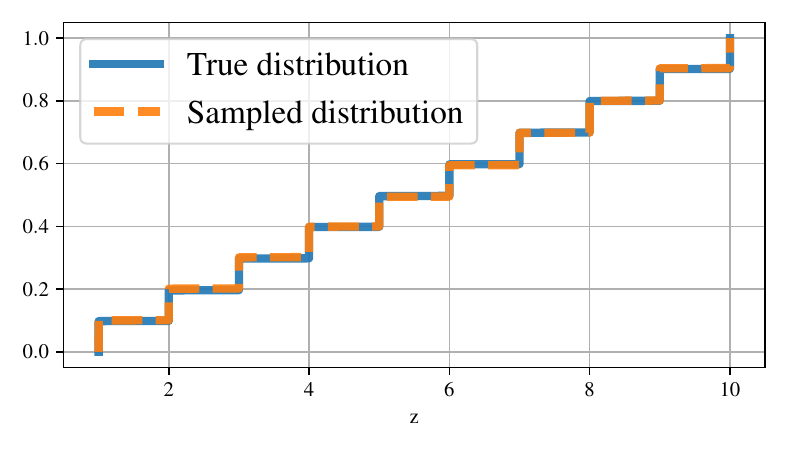}}
        \vspace{-5mm}
        \caption{Discrete, $\ctofarthree$ with $20+20+20$ bins}
      \end{subfigure}
  }}
  \caption{Distribution recovery supplemental results on the Gaussian
    mixture model synthetic data (GMM, left) and the discrete uniform
    synthetic data (Discrete, right) for each of our implemented
    systems.\label{fig:supp_recovery}}
\end{figure}

Fig.~\ref{fig:supp_recovery} shows the ability of each of our implemented
systems to recover the synthetic distribution (repeating the
$\ctofarthree$ results, also given in Fig.~5 in the main paper).
The frequently-used Gaussian output
distribution~\cite{li2019enhancing,salinas2020deepar,chen2020probabilistic}
cannot recover the Gaussian mixture since it has only a single mixture
component, while it fits the discrete data as well as can be expected.
For the $\ctofarRNN$ models, we use 60 total \emph{bins} across all
the levels, which amounts to much greater precision for the
multi-level models.
Qualitatively, $\ctofarone$ fits the data fairly well, not quite
overlapping the true GMM distribution, and struggling to generate the
sharp increases in the CDF seen in the discrete data.  The
$\ctofartwo$ and $\ctofarthree$ models fit each distribution nearly
perfectly.

{\begin{table}
\caption{Negative log likelihood (NLL) of test data for different
  models on synthetic data\label{tab:dist_recovery}}
\footnotesize
\begin{tabular}{@{}lll@{}}
\toprule
                  & GMM      & Discrete \\ \midrule
$\deepargaussian$ & 0.2526 & 0.2775 \\
$\ctofarone$      & -0.4150 & -1.564 \\
$\ctofartwo$      & -0.4198 & -4.479 \\
$\ctofarthree$    & -0.4203 & -6.664 \\ \bottomrule
\end{tabular}
\end{table}
}

Quantitatively, we can evaluate each of these fits by computing the
negative log likelihood (NLL) of the test data, as computed by each
model.  These results are presented in Table~\ref{tab:dist_recovery},
which shows $\ctofarthree$ is only very slightly better on GMM, but
significantly better on the discrete data.  Of course, by adding more
layers and bins, we may achieve arbitrarily low NLL on the discrete
dataset.  Whether such precision is beneficial will depend on the
application.

\section{Empirical study on real-world data: further details}

\subsection{Azure VM demand dataset}\label{subsec:azure}

{\begin{table}
\caption{Information about the $\azure$ dataset.  Flavor \emph{names}
  are automatically generated from each VM's allocated
  \emph{VCPUs} and \emph{Memory} requirement via the template
  \emph{az-\{VCPUs\}-\{Memory/VCPUs\}}.  Flavor \emph{types} are
  defined purely by the Memory:VCPUs ratio via the template
  \emph{az-\{Memory/VCPUs\}}.\label{tab:azure}}
\footnotesize
\begin{tabular}{@{}ll@{}}
\toprule
Duration                  & 30 days
\\
Num\@. unique VMs           & 2,013,767
\\
Num\@. unique subscriptions & 5,958, with top 250 used for customer-specific series.
\\
Unique VM categories      & \{Delay-insensitive, Interactive,
Unknown\} \\
Unique flavor names &
  \begin{tabular}[c]{@{}l@{}}\{az-1-1.75, az-1-2, az-16-7, az-2-1.75, az-2-2, az-2-7, az-2-8, az-4-1.75,\\   az-4-2, az-4-7,
    az-4-8,  az-8-1.75, az-8-2, az-8-7, az-8-8\}\end{tabular} \\
Unique flavor types       & \{az-0.75, az-1.75, az-2, az-7,   az-8\}
\\ \bottomrule
\end{tabular}
\end{table}
}

We generated the $\azure$ dataset in order to provide real-world data
reflecting the types of time series seen in the context of large-scale
compute clouds.
Cloud decision making can benefit from predicted future resource
demand, e.g., for capacity planning or scheduling
optimization~\cite{bergsma2021generating}.
A dominant workload type in compute clouds is the virtual machine
(VM), which is typically available in one of a limited number of
specific configurations or \emph{flavors}; essentially, a flavor
represents a specific bundle of resource requirements, including VCPU
and Memory needs.
Forecasts for the total demand (in terms of VCPUs or Memory, in GB) of
specific flavors, customers, and workload categories are all valuable.
Furthermore, aggregations of these basic time series are also
valuable.  For example, VM flavors with a common VCPUs:Memory ratio,
known as a VM flavor \emph{type}, are often run on shared physical
servers, and therefore forecasts of flavor type demand are directly
actionable in terms of provisioning of server resources.

We obtained real-world data reflecting these dimensions of cloud
resource demand by leveraging the publicly-available \emph{Azure
  Public
  Dataset}\footnote{\url{https://github.com/Azure/AzurePublicDataset/blob/master/AzurePublicDatasetV1.md}},
originally released in~\cite{cortez2017resource} under a Creative
Commons Attribution 4.0 International Public License.
This dataset contains create, delete, and CPU utilization information
(as a \% of allocated VCPUs, over time) for over 2 million cloud VMs,
reflecting both internal and external customer workload over a 30-day
period.
We converted this data into time series by counting the total VCPUs
and Memory requirements over time for different combinations of
flavors, subscriptions (customer IDs), categories, and flavor types,
as noted in Table~\ref{tab:azure}.
Although we count VMs from all customers, we build customer-specific
time series for the top 250 subscription IDs by VM frequency.
We also combined lower-level time series to form a hierarchy of time
series, as is common in real-world demand
data~\cite{taieb2017coherent,rangapuram2021end}.
We also defined a \emph{stopped} VM as any VM whose CPU utilization\%
drops below 1\%.
We then created two different versions of our time series, in each
case aggregating the data with a 1-hour sampling period:
\begin{enumerate}
\item The time series aggregated with the \emph{maximum} of each
  1-hour period.
\item The time series aggregated with the \emph{minimum} of each
  1-hour period, and excluding stopped VMs at each time point.
\end{enumerate}
In a way, the first set of time series represents a \emph{pessimistic}
view of how many resources we require each hour, while the second
version represents an \emph{optimistic} view, as we only provision for
the minimum and assume we can re-use the resources from stopped VMs.

\begin{figure}
  \centering
      {\includegraphics[width=1.0\textwidth]{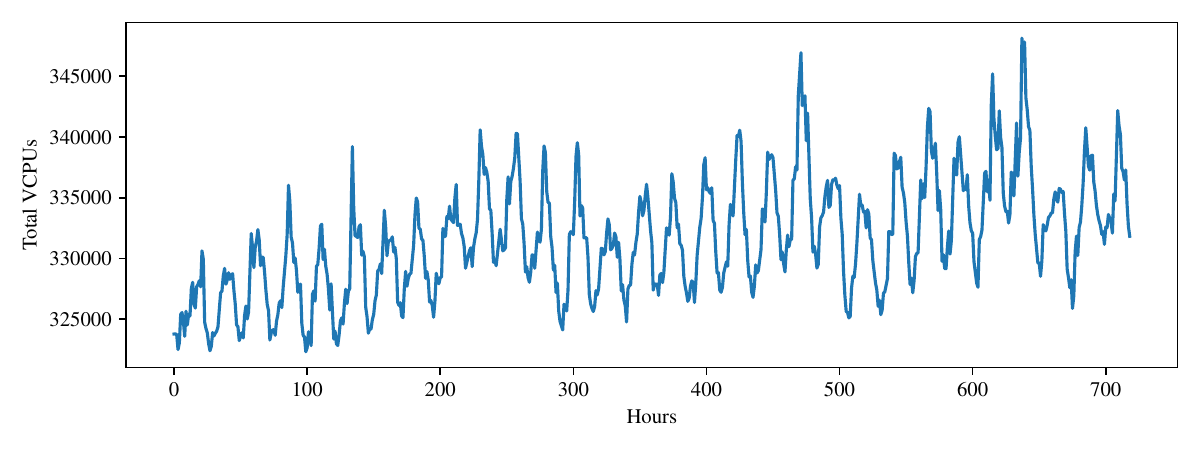}}
      \caption{$\azure$ sample time series: {total} \emph{aggregate} VCPUs demand.
        A strong daily seasonality is apparent.\label{fig:azure_all}}
\end{figure}

\begin{figure}
  \centering
      {\includegraphics[width=1.0\textwidth]{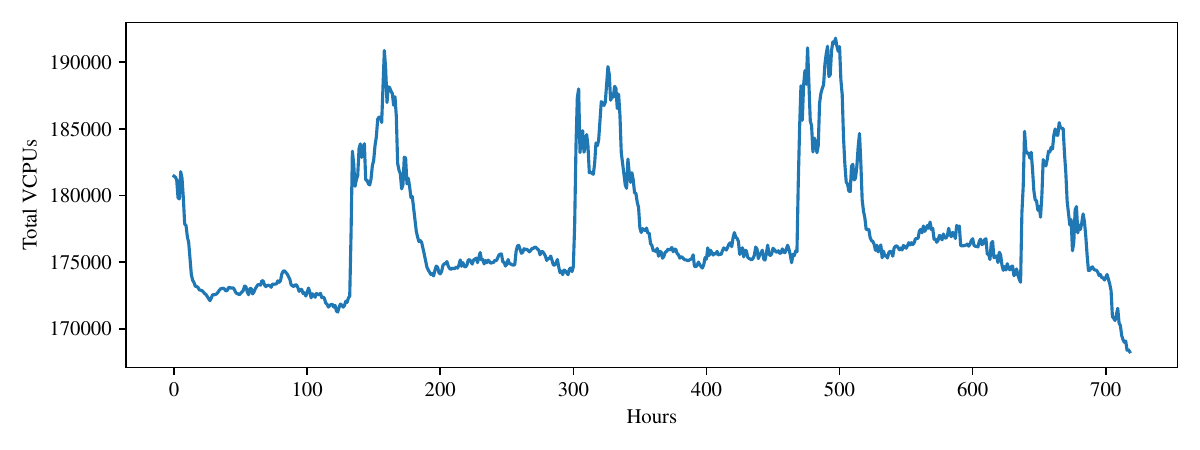}}
      \caption{$\azure$ sample time series: total VCPUs demand of
        \emph{delay-insensitive} VMs of flavor type \emph{az-1.75}.
        \emph{Delay-insensitive} VMs exhibit strong weekly seasonality,
        apparently being used mostly on weekends.\label{fig:azure_delayinsensitive}}
\end{figure}

\begin{figure}
  \centering
      {\includegraphics[width=1.0\textwidth]{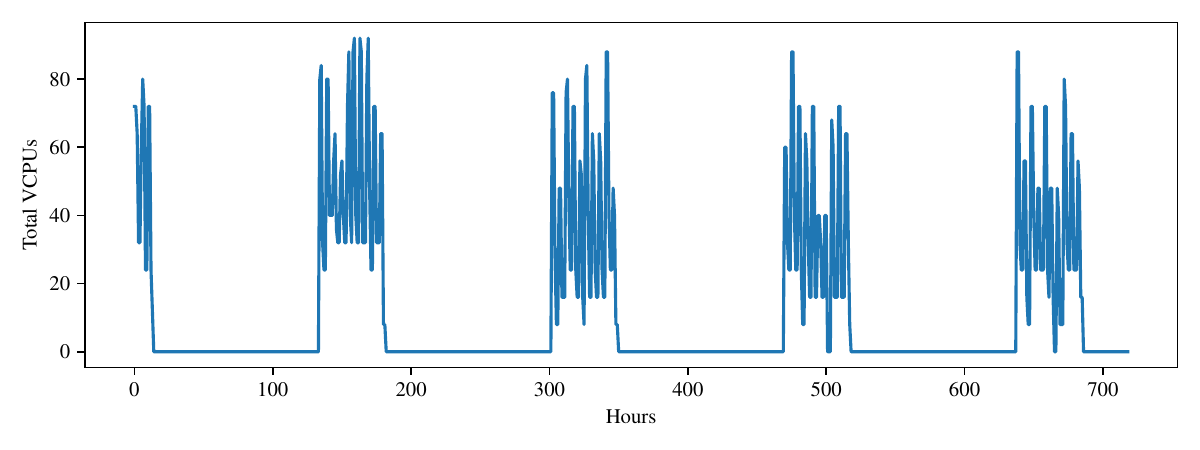}}
      \caption{$\azure$ sample time series: customer 1108, total VCPUs
        demand of \emph{delay-insensitive} VMs of flavor type az-1.75.
        This particular customer seems to consume resources
        exclusively on weekends.\label{fig:azure_cust1108}}
\end{figure}

\begin{figure}
  \centering
      {\includegraphics[width=1.0\textwidth]{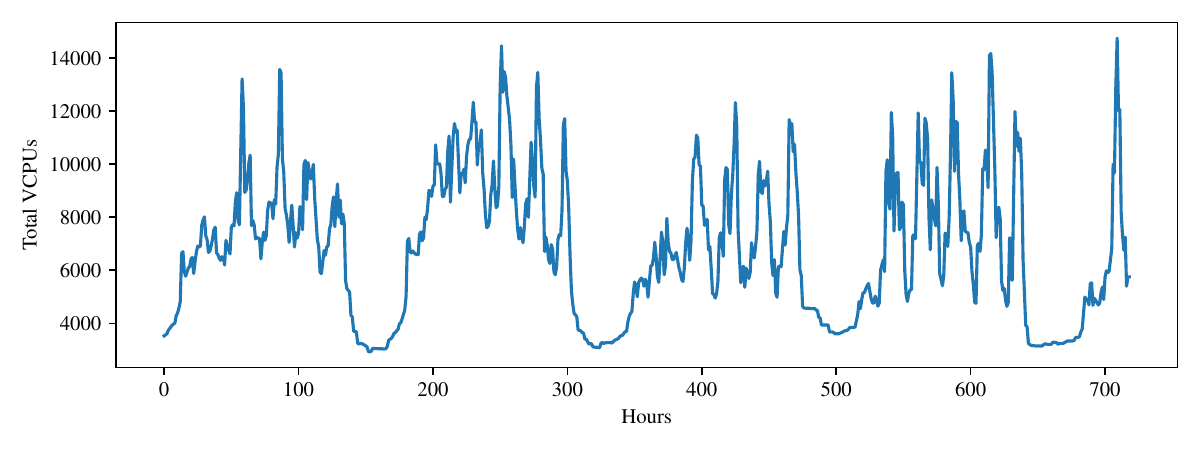}}
      \caption{$\azure$ sample time series: total VCPUs demand of
        \emph{unknown} VMs of flavor name az-8-1.75.  \emph{Unknown}
        VMs exhibit strong daily and weekly seasonality, requiring
        resources during the periods where \emph{delay-insensitive} VMs
        are lower (i.e., during weekdays).\label{fig:azure_unknown}}
\end{figure}

We provide some examples of these time series in
Figs.~\ref{fig:azure_all} through~\ref{fig:azure_unknown}.
Fig.~\ref{fig:azure_all} represents the total demand across all VMs,
and we can see the strong daily seasonality.
Demand of the \emph{delay-insensitive} category, for a particular
flavor type, is shown in Fig.~\ref{fig:azure_delayinsensitive}, while
Fig.~\ref{fig:azure_cust1108} gives the demand for the same category
and flavor type, but for customer 1108 specifically (note all the
original IDs are anonymized and represented in the dataset using
placeholder values).
Demand of the \emph{unknown} category, for flavor name az-8-1.75, is given in
Fig.~\ref{fig:azure_unknown}.

Although the $\azure$ dataset covers 30 days, for experiments we use
20 days as training, 3 days for validation, and 3 final days for
testing, leaving the remaining few days unseen and available for
future experiments.

\subsection{Other datasets and dataset details}

{\begin{table}
\caption{Dataset details for the empirical study.\label{tab:datasets}}
\footnotesize
\begin{tabular}{@{}llllllllll@{}}
\toprule
Dataset &
  \begin{tabular}[c]{@{}l@{}}Binning\\ extent\\ start\end{tabular} &
  \begin{tabular}[c]{@{}l@{}}Binning\\ extent\\ end\end{tabular} &
  \begin{tabular}[c]{@{}l@{}}Num.\\ series\end{tabular} &
  Domain &
  Freq. &
  \begin{tabular}[c]{@{}l@{}}Num.\\ vals per\\ series\end{tabular} &
  \begin{tabular}[c]{@{}l@{}}Num.\\ validation\\ vals per\\ series\end{tabular} &
  \begin{tabular}[c]{@{}l@{}}Num.\\ test vals\\ per series\end{tabular}
    &
  \begin{tabular}[c]{@{}l@{}}Prediction\\ range size\end{tabular}
    \\ \midrule
$\elec$ & -0.01 & 1.01 & 321  & Discrete & Hourly & 21212 & 168 &
168 & 24 \\
$\traffic$     & -0.01 & 1.01 & 862  & Real   & Hourly & 14204 & 168 &
168 & 24 \\
$\wiki$        & -0.2  & 3.1  & 9535 & Discrete & Daily  & 912   & 50  &
150 & 30 \\
$\azure$       & -0.1  & 1.3  & 4098 & Discrete & Hourly & 719   & 72  &
72  & 24 \\ \bottomrule
\end{tabular}
\end{table}
}

Beyond $\azure$, the $\elec$, $\traffic$, and $\wiki$ datasets were
obtained using scripts in GluonTS~\cite{alexandrov2020gluonts}, with
the objective of replicating the training/validation/test splits used
in prior
work~\cite{salinas2019high,rabanser2020effectiveness,gouttes2021probabilistic}.
Table~\ref{tab:datasets} provides the details of these datasets and
$\azure$.  Note the binning extent was selected in order to cover from
roughly the 1\% to the 99\% percentiles of normalized values in the
training data for each series (normalized using min-max scaling on the
\emph{conditioning} range; the prediction range can go below the min
and above the max).

\subsection{Metrics}

Let $i$ index the time series, and $t$ index the time step.  Let $N$
be the total number of points $z_{i,t}$ on which we evaluate.  Let
$\mathcal{I}(\cdot)$ denote the indicator function. Let
$\hat{z}_{i,t}^{(q_\#)}$ be the $q_\#$ quantile of the forecast
distribution for time series $i$ at point $t$,
e.g. $\hat{z}_{i,t}^{(q_{0.8})}$ is the value such that 80\% of
possible values for point $z_{i,t}$ are expected to be below this
value.

We define \emph{pinball loss} and \emph{quantile loss} as part of the
derivation of \emph{weighted quantile loss}.  \emph{Weighted quantile
  loss}, \emph{normalized deviation}, \emph{calibration}, and
\emph{sharpness} are reported in the main paper.

\paragraph{Pinball loss:}

{\begin{equation*}
\Lambda_\alpha(q,z)=(\alpha - \mathcal{I}(z < q))(z-q)
\end{equation*}
}

\paragraph{Quantile loss:}

{\begin{equation*}
\text{QL\#} = \frac{\sum_{i,t} 2\Lambda_{(q_\#)}(\hat{z}_{i,t}^{(q_\#)}, z_{i,t})}{\sum_{i,t}|z_{i,t}|}
\end{equation*}
}

\paragraph{Weighted quantile loss:}

{\begin{equation*}
\text{wQL} = \frac{1}{9}(\text{QL0.1} + \text{QL0.2} + \dots + \text{QL0.9})
\end{equation*}
}

\paragraph{Normalized deviation:}

{\begin{equation*}
\text{ND} = \frac{\sum_{i,t}|z_{i,t} - \hat{z}_{i,t}|}{\sum_{i,t}|z_{i,t}|}
\end{equation*}
}

\paragraph{Calibration:}

{\begin{equation*}
\text{calibration} = \frac{\sum_{i,t} \mathcal{I}(\hat{z}^{(q_l)}_{i,t} < z_{i,t} \le \hat{z}^{(q_u)}_{i,t})}{N}
\end{equation*}
}

\paragraph{Sharpness:}

{\begin{equation*}
\text{sharpness} = \frac{\sum_{i,t} |\hat{z}^{(q_u)}_{i,t} - \hat{z}^{(q_l)}_{i,t}|}{\sum_{i,t}|z_{i,t}|}
\end{equation*}
}

Where calibration and sharpness are reported together as the CovX
metric.  E.g., Cov80 gives the coverage and sharpness of the 80\%
prediction range where $q_l = q_{0.1}$ and $q_u = q_{0.9}$.
Note that for all metrics we use 500 Monte Carlo samples in order to
generate our forecast distribution, except for generating the Cov99
metrics in Table~1 of the main paper, for which we used 5000 samples,
as the extreme percentiles are by definition more rare and require
more samples for good estimation.

\subsection{Background and motivation for our experimental setup}

As mentioned in the main paper, multi-level $\ctofar$ models actually
include shallower models as special cases (i.e., a three-level
$\ctofar$ model is equivalent to a one-level model with a single bin
in both the second and the third levels).  So given enough tuning,
$\ctofar$ is strictly more powerful than a flat binning.
Moreover, any shallower model (including a flat binning) could
potentially be improved by adding additional finer-grained $\ctofar$
levels, increasing the precision without making the problem any more
complex for the original shallower model.
We initially used this approach, adding a second $\ctofar$ level to
our production flat binning system and finding gains across all our
internal and publicly-available datasets.
In essence, the finer-grained $\ctofar$ level acts as a kind of
``reconstruction function''~\cite{rabanser2020effectiveness} for the
coarse, flat binning model; the lower-level network maps the top-level
bin to a more precise location.  Adding another lower-level model
increases the precision further, again, without any cost to the
coarser predictions, and further levels can be added recursively until
the optimum level of precision is obtained for the task at hand.

However, adding additional levels does have two practical costs: each
level increases both the number of parameters (via the added networks
at each level) and the number of hyperparameters (that is, the number
of bins at each new level).
Based on both theory and initial experimental results, we are
therefore confident that we can pay this cost in order to achieve
superior forecasting accuracy.
However, for the purposes of this paper, we elected to pursue a
different experimental question: for a fixed parameter budget
(counting parameters across all levels in the $\ctofar$ hierarchy),
can we jointly tune the number of bins at each level in order to
achieve superior predictions compared to traditional approaches?  And
can we do this without any additional cost for hyperparameter tuning?
To answer this question, we established a fixed parameter budget of
one million parameters for each system, and a budget of 100 tuning
trials for each system.

\subsection{Tuned and fixed hyperparameters}\label{sec:tuning_setup}

{\begin{table}
\caption{Tuning ranges for hyperparameter optimization.\label{tab:tuning}} \footnotesize
\begin{tabular}{@{}lll@{}}
\toprule
Hyperparameter                                    & Range            &Sampling type \\ \midrule
n\_hidden                                   & {[}16, 288{]}     &integer       \\
learning\_rate                                    & {[}1e-5, 1e-1{]} &loguniform    \\
weight\_decay                                     & {[}1e-7, 1e-2{]} &loguniform    \\
n\_bins, for $\ctofarone$                             & {[}4, 1024{]}    &integer       \\
n\_bins\_level\_$i$, $i$th level of $\ctofarB$, $B > 1$ & {[}4, 128{]}     &integer       \\ \bottomrule
\end{tabular}
\end{table}
}
{\begin{table}
  \caption{Fixed hyperparameters used in the empirical study.\label{tab:fixed}}
  \scriptsize
\begin{tabular}{@{}lll@{}}
\toprule
Hyperparameter                                       & Value     &
Note
\\ \midrule
n\_max\_total\_parms                                 & 1,000,000 &
Enforced via a cap on n\_hidden                                \\
lstm\_dropout                                        & 1e-3      &
Intra-layer dropout                                         \\
n\_lstm\_layers                                      & 2         &
\\
n\_conditioning\_range-hourly & 168       &  For $\elec$, $\traffic$, $\azure$, as in prior work                                            \\
n\_conditioning\_range-daily                  & 30        & For $\wiki$, as in prior work                                            \\
n\_rollouts (validation)                             & 25        & For computing ND on validation set                          \\
n\_rollouts (test)                                   & 500       & For evaluation on test set                                  \\
n\_train\_batch\_size                                & 1024      & Total num\@. prediction ranges per batch                 \\
n\_train\_ranges\_per\_checkpoint                          & 32768 & Total num\@. prediction ranges in one \emph{checkpoint} (train loss reported)        \\
n\_validation\_set                                   & 32768     & Total num\@. prediction ranges per validation evaluation \\
n\_max\_checkpoints                                  & 750       & Maximum num\@. checkpoints (sets of n\_train\_ranges\_per\_checkpoint)       \\
n\_validation\_eval\_period                          & 2         & Num\@. train checkpoints per validation evaluation       \\
n\_validation\_eval\_warmup & 11 & Num\@. train checkpoints before first validation evaluation       \\
n\_stop\_evals\_no\_improve & 37 & Num\@. validation evaluations without improvement before stopping
\\ \bottomrule
\end{tabular}
\end{table}
}

The specific parameters that are tuned are given in
Table~\ref{tab:tuning}.
Given tuning search space grows exponentially with added
hyperparameters, we elected to fix some hyperparameters to values that
proved effective in preliminary experiments (e.g.,
\emph{lstm\_dropout} and \emph{n\_lstm\_layers}); other
hyperparameters are either set to follow prior work (e.g., the
conditioning range parameters), or are set for practical reasons in
order to help maximize the use of our available compute resources.
See Table~\ref{tab:fixed} for a list of all fixed hyperparameters.

Recall that for forecasting, we tune for multi-step-ahead normalized
deviation (ND) on the validation set.  Whether tuning $\ctofar$ models
or tuning $\deepargaussian$, this requires running the Monte Carlo
sampling procedure to generate a forecast distribution; we use the
median of this forecast distribution as the point forecast for
evaluation.  Since sampling is relatively expensive, we evaluate only
every second checkpoint, only after 11 warm-up checkpoints, on only
32768 validation prediction ranges, and only using 25 rollouts to
generate the forecast distribution (note corresponding hyperparameters
in Table~\ref{tab:fixed}).  We also use smaller batch sizes for
generating forecasts (testing) than we do in training (since we
simultaneously vectorize over Monte Carlo rollouts for each series).

\subsection{Tuning results}

{\begin{table}
\caption{Tuning results, tuning for normalized deviation, in the
  empirical study.\label{tab:tuning_results}} \footnotesize
\begin{tabular}{@{}llllllll@{}}
\toprule
Dataset     & System            & nhidden & NBins1 & NBins2 & NBins3 &
Total bins & Total intervals \\ \midrule
$\elec$ & $\deepargaussian$ & 141     & -      & -      & -      & - & -               \\
$\elec$ & $\ctofarone$      & 248     & 110    & -      & -      & 110        & 110             \\
$\elec$ & $\ctofartwo$      & 189     & 12     & 35     & -      & 47         & 420             \\
$\elec$ & $\ctofarthree$    & 156     & 5      & 25     & 21     & 51         & 2625            \\ \midrule
$\traffic$     & $\deepargaussian$ & 165     & -      & -      & -      & -& -               \\
$\traffic$     & $\ctofarone$      & 236     & 163    & -      & -      &163        & 163             \\
$\traffic$     & $\ctofartwo$      & 184     & 28     & 13     & -      &41         & 364             \\
$\traffic$     & $\ctofarthree$    & 146     & 17     & 9      & 5      &31         & 765             \\ \midrule
$\wiki$        & $\deepargaussian$ & 153     & -      & -      & -      & -& -               \\
$\wiki$        & $\ctofarone$      & 146     & 968    & -      & -      &968        & 968             \\
$\wiki$        & $\ctofartwo$      & 176     & 21     & 18     & -      &39         & 378             \\
$\wiki$        & $\ctofarthree$    & 139     & 79     & 16     & 11     &106        & 13904           \\ \midrule
$\azure$       & $\deepargaussian$ & 249     & -      & -      & -      & -& -               \\
$\azure$       & $\ctofarone$      & 83      & 75     & -      & -      &75         & 75              \\
$\azure$       & $\ctofartwo$      & 64      & 16     & 71     & -      &87         & 1136            \\
$\azure$       & $\ctofarthree$    & 130     & 16     & 11     & 94     &121        & 16544          \\ \bottomrule
\end{tabular}
\end{table}
}

Table~\ref{tab:tuning_results} provides the results of our tuning
procedure in terms of the selected number of hidden units and bins.
While prior work used a fixed 1024 bins in their flat
binning~\cite{rabanser2020effectiveness}, our tuner often selected
quite fewer bins for $\ctofarone$ on our datasets.  $\ctofartwo$ and
$\ctofarthree$ models generally use fewer total bins than flat
binning, while having very many more total actual intervals.

\subsection{Computational performance and resource requirements}

{\begin{table}
\caption{Training time in hours for top system on validation set in
  empirical study.\label{tab:training_time}}
\footnotesize
\begin{tabular}{@{}llllll@{}}
\toprule
                  & $\elec$ & $\traffic$ & $\wiki$ & $\azure$ & Average
\\ \midrule
$\deepargaussian$ & 18.2 & 31.3    & 1.0  & 5.0   & 13.9    \\
$\ctofarone$      & 28.7 & 27.1    & 12.8 & 12.8  & 20.3    \\
$\ctofartwo$      & 70.0 & 72.1    & 22.5 & 6.8   & 42.9    \\
$\ctofarthree$    & 29.5 & 57.1    & 32.3 & 31.5  & 37.6
\\ \bottomrule
\end{tabular}
\end{table}
}

{\begin{table}
\caption{Time per 100 forecasts in \emph{seconds} (running on NVIDIA
  Tesla P100) by top system on test set. All systems ran with common
  test batch sizes (60 for daily $\wiki$ dataset, 22 for hourly
  datasets) and number of samples (500) for each
  dataset.\label{tab:testing_time}} \footnotesize
\begin{tabular}{@{}llllll@{}}
\toprule
                  & $\elec$ & $\traffic$ & $\wiki$ & $\azure$ & Average
\\ \midrule
$\deepargaussian$ & 1.28 & 1.57    & 0.55 & 2.72 & 1.53    \\
$\ctofarone$      & 3.73 & 3.79    & 1.71 & 1.34 & 2.64    \\
$\ctofartwo$      & 4.22 & 4.12    & 1.50 & 1.68 & 2.88    \\
$\ctofarthree$    & 4.81 & 4.45    & 1.92 & 4.73 & 3.98   
\\ \bottomrule
\end{tabular}
\end{table}
}

{\begin{table}
\caption{Amount of memory consumed for prediction in MiB (measured via
  \emph{nvidia-smi} on NVIDIA Tesla P100) by top system on test set.
  All systems ran with common \emph{test batch sizes} (60 for daily
  $\wiki$, 22 for hourly datasets) and number of samples (500) for
  each dataset.  The flat binning, $\ctofarone$, consumes
  significantly more memory on three of the four datasets; memory
  requirements depend directly on the number of bins selected by the
  tuner (see Table~\ref{tab:tuning_results}).\label{tab:resources}}
\footnotesize
\begin{tabular}{@{}llllll@{}}
\toprule
                  & $\elec$ & $\traffic$ & $\wiki$  & $\azure$ & Average
\\ \midrule
$\deepargaussian$ & 4295 & 4873    & 3113  & 4845  & 4281.50 \\
$\ctofarone$      & 8264 & 7465    & 12619 & 3389  & 7934.25 \\
$\ctofartwo$      & 6233 & 6147    & 5577  & 4209  & 5541.50 \\
$\ctofarthree$    & 5657 & 5111    & 6372  & 6037  & 5794.25
\\ \bottomrule
\end{tabular}
\end{table}
}

Table~\ref{tab:training_time} has the training times for the top
systems found on the validation set.  Training time naturally reflects
both the speed of convergence in learning (number of training epochs)
\emph{and} the speed of operating the specific
architecture.\footnote{Note that the baseline systems $\naive$,
  $\seasonalnaive$, and $\ets$ do not require training; $\ets$
  parameters are fit separately for each input history at
  \emph{inference} time.}

Testing time roughly follows a similar pattern
(Table~\ref{tab:testing_time}), taking longer on the multi-level
$\ctofar$ models, although more efficient implementations than ours
are certainly possible.
Meanwhile, Table~\ref{tab:resources} gives the memory requirements of
the different systems.
Overall, we may say that in our implementation, multi-level $\ctofar$
models run slower than a flat binning, but with less memory.

\subsection{Evaluation by forecast horizon}\label{subsec:horizons}

\begin{figure}
  \vspace{0mm}
  \centering {\makebox[\textwidth][c]{
      \hspace{-3mm}
      \begin{subfigure}{\shrinkfigtwo\textwidth}
        \begin{tikzpicture}
          \node (img1) {\includegraphics[width=\textwidth]{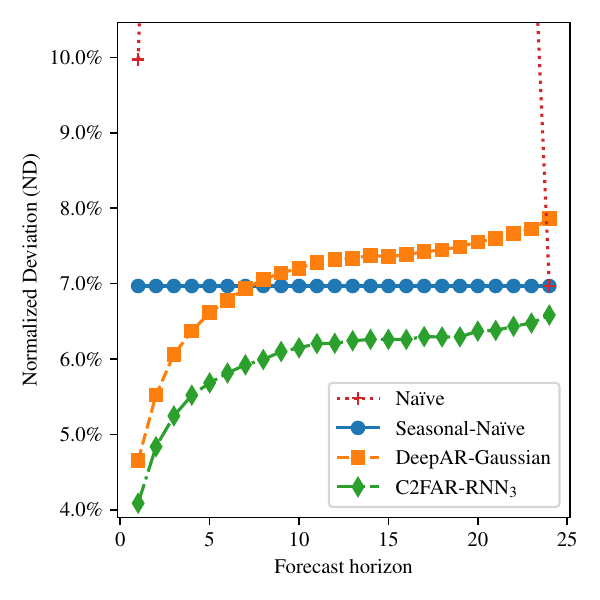}};
          \node[below of=img1, fill=lime, align=center, yshift=2.1cm, xshift=-1.9cm, node distance=0cm, anchor=west] {$\elec$};
        \end{tikzpicture}
      \end{subfigure}
      \hspace{-1mm}
      \begin{subfigure}{\shrinkfigtwo\textwidth}
        \begin{tikzpicture}
          \node (img1) {\includegraphics[width=\textwidth]{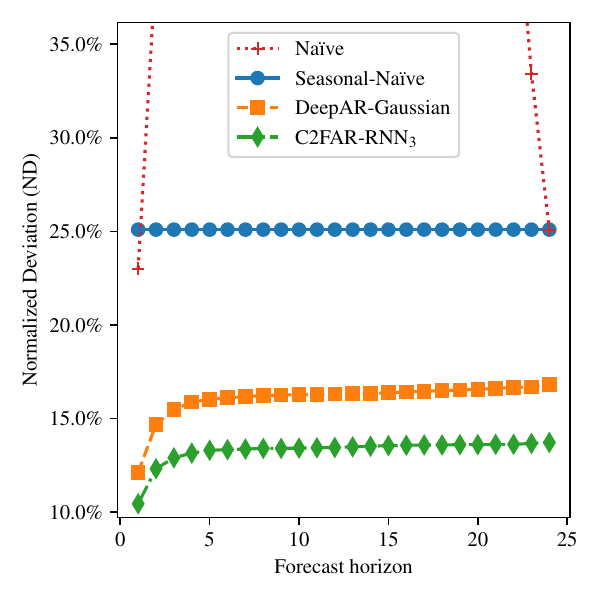}};
          \node[below of=img1, fill=lime, align=center, yshift=2.1cm, xshift=-1.9cm, node distance=0cm, anchor=west] {$\traffic$};
        \end{tikzpicture}
      \end{subfigure}
  }} \\
  \vspace{-3mm}
  {\makebox[\textwidth][c]{
      \hspace{-3mm}
      \begin{subfigure}{\shrinkfigtwo\textwidth}
        \begin{tikzpicture}
          \node (img1) {\includegraphics[width=\textwidth]{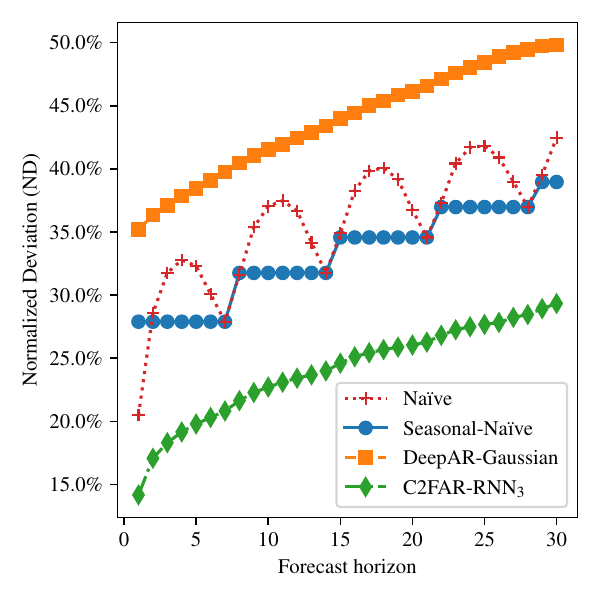}};
          \node[below of=img1, fill=lime, align=center, yshift=2.1cm, xshift=-1.9cm, node distance=0cm, anchor=west] {$\wiki$};
        \end{tikzpicture}
      \end{subfigure}
      \hspace{-1mm}
      \begin{subfigure}{\shrinkfigtwo\textwidth}
        \begin{tikzpicture}
          \node (img1) {\includegraphics[width=\textwidth]{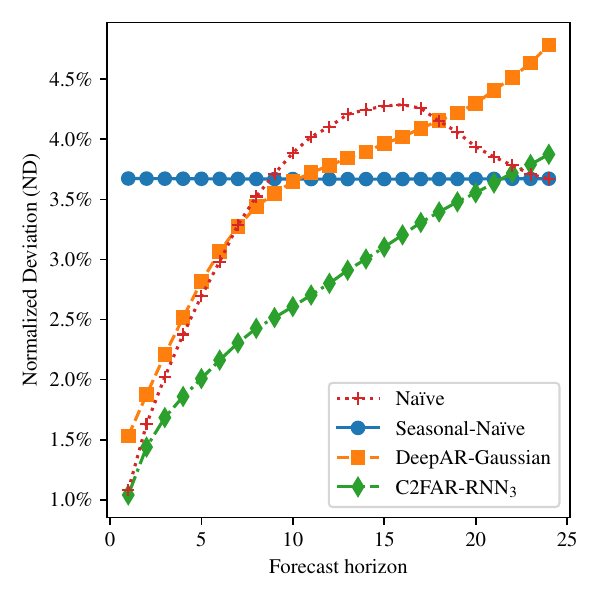}};
          \node[below of=img1, fill=lime, align=center, yshift=2.1cm, xshift=-1.9cm, node distance=0cm, anchor=west] {$\azure$};
        \end{tikzpicture}
      \end{subfigure}
  }}
  \vspace{0mm}
  \caption{Comparison of different forecasting systems with normalized
    deviation (ND) calculated separately at each forecast horizon.
    For $\elec$, $\traffic$, and $\azure$, we forecast forward for one
    24-hour seasonal cycle, while for $\wiki$, we predict for
    slightly-more-than-four 7-day cycles.  $\seasonalnaive$ is flat
    over a cycle because we evaluate using rolling predictions: every
    datapoint is forecast once at every horizon, and always gets the
    same prediction.  Vanilla $\naive$ becomes first less accurate,
    then more accurate as we approach the end of the cycle, at which
    point it becomes equivalent to $\seasonalnaive$.  Aside from
    $\wiki$, where $\deepargaussian$ fails to learn a good model,
    $\deepargaussian$ is competitive with $\ctofarthree$ at earlier
    horizons, but the gap widens over time.\label{fig:horizs}}
\end{figure}
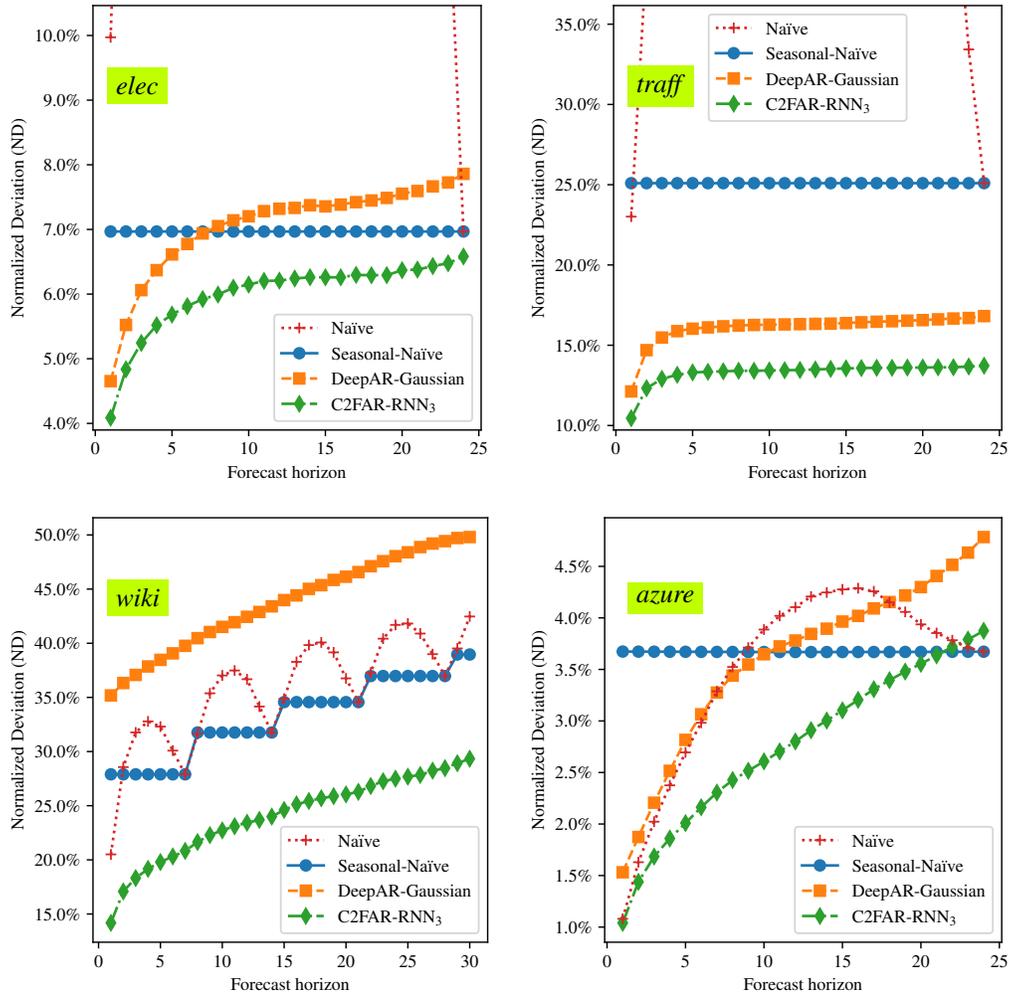

Figure~\ref{fig:horizs} shows the forecast error of the systems as a
function of the forecast horizon.  Compared to three baselines
($\naive$, $\seasonalnaive$ and $\deepargaussian$), we find
$\ctofarthree$ is the best performer across virtually all horizons on
all datasets, except the last couple horizons on $\azure$.
Interestingly, before $\ctofar$, practictioners faced a very nuanced
problem when selecting a forecast method for their own dataset.  E.g.,
on the cloud demand data that we are most interested in ($\azure$),
and ignoring $\ctofar$, it seems that on some horizons, $\naive$ is
best, on others, DeepAR, and on others still, it is $\seasonalnaive$.
$\ctofar$ combines the best aspects of all of these systems,
outputting highly-precise predictions when doing so makes sense
(similar to $\naive$), but generating seasonally-adjusted estimates
during the middle forecast horizons.

\section{Stability of Empirical Results}

In this section, we investigate the stability of our empirical
results.
Random seeds are used in both our testing process (via Monte Carlo
sampling of predicted future values) and our tuning process (via
sampling of hyperparameters), and it is important to quantify the
stability of these sources of randomness
separately~\cite{clark2011better}.
Ideally, we would repeat our entire tuning procedure
multiple times with different random seeds, allowing us to determine
the reliability of our process for fitting both model parameters and
hyperparameters.
While such repetition is not practical given the total time required,
we can nevertheless investigate tuning randomness in other ways, and
use this to assess the stability of our empirical results.
Note that stability of results may depend on both the systems under
investigation (e.g., models with more hyperparameters may be more
unstable with respect to tuning), the datasets and splits used in the
experiments (e.g., larger data sets may be more stable), and the
tuning/training/testing processes (e.g., more tuning runs may lead to
more stable results).

We summarize the results as follows, but provide full details in the
following subsections:
\begin{itemize}
\item The Monte Carlo sampling process is very stable with respect to
  the random seed: we repeated the sampling 6 times and found negligible
  differences in normalized deviation (\S\ref{subsec:sampling}).
\item Results are fairly stable when moving from the validation set to
  the test set: on all validation sets and all test sets, the top
  $\ctofar$ models improve on the top $\deepargaussian$ model.  On all
  validation sets and all test sets, $\ctofartwo$ performed better
  than $\ctofarone$.  However, in testing three of the four datasets,
  $\ctofarthree$ improved over its ranking on the validation set
  (\S\ref{subsec:valtest}).
\item Results are less stable with respect to the tuning.  If we use
  the second-best validation-set model on each \emph{test} set,
  $\ctofarthree$ and $\ctofartwo$ suffer larger drops on test than
  $\ctofarone$, while $\deepargaussian$ improves in two cases.
  However, even with the second-best models, $\ctofar$ models perform
  better than $\deepargaussian$ across all test sets, and
  $\ctofarthree$ remains the top system on three of the four test sets
  (\S\ref{subsec:tuning}).
\end{itemize}

Note we do not assess the stability of re-training the models with the
same hyperparameters, but different shuffling of the training data, as
such re-training is not currently part of our operations.  However, as
future work, we plan to investigate the stability of
re-training-without-re-tuning in the context of slight increases in
the training data over time, which is very common in real-world ML
pipelines.

\subsection{Sampling stability}\label{subsec:sampling}

\begin{figure}
  \centering
      {\includegraphics[width=1.0\textwidth]{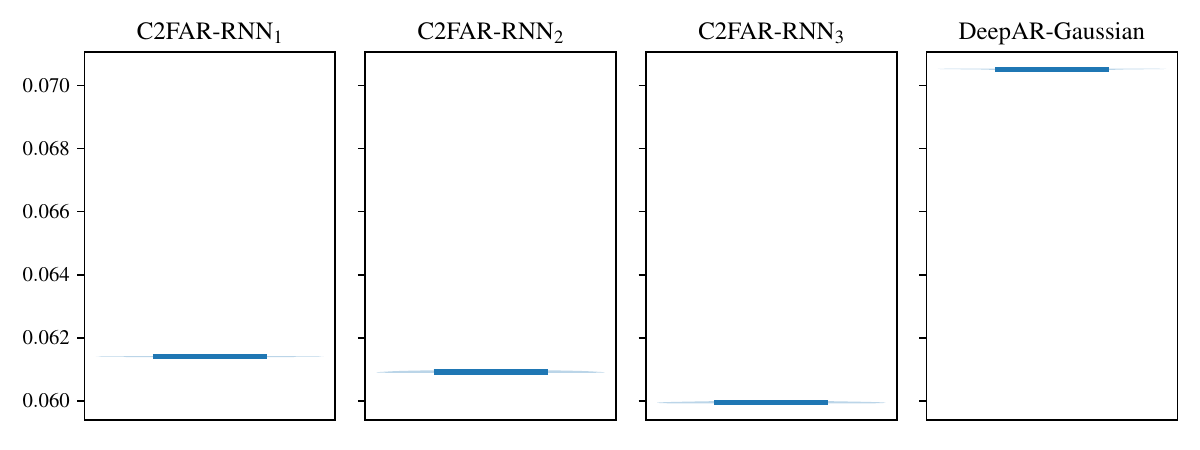}}
      \caption{$\elec$ sampling stability: distribution of normalized
        deviation (on test set) across different random seeds as a violin plot.\label{fig:sampling_elec}}
\end{figure}

\begin{figure}
  \centering
      {\includegraphics[width=1.0\textwidth]{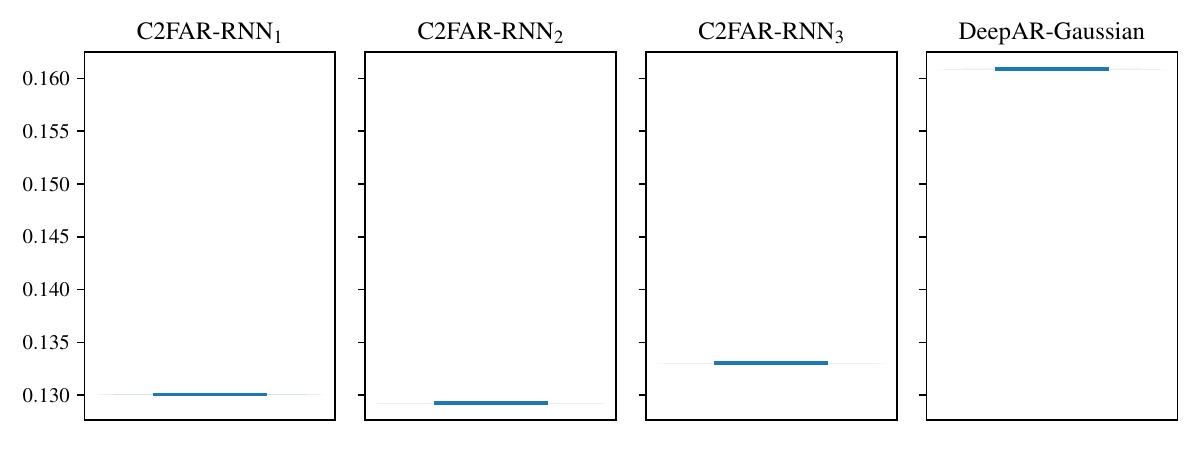}}
      \caption{$\traffic$ sampling stability: distribution of normalized
        deviation (on test set) across different random seeds as a violin plot.\label{fig:sampling_traffic}}
\end{figure}

\begin{figure}
  \centering
      {\includegraphics[width=1.0\textwidth]{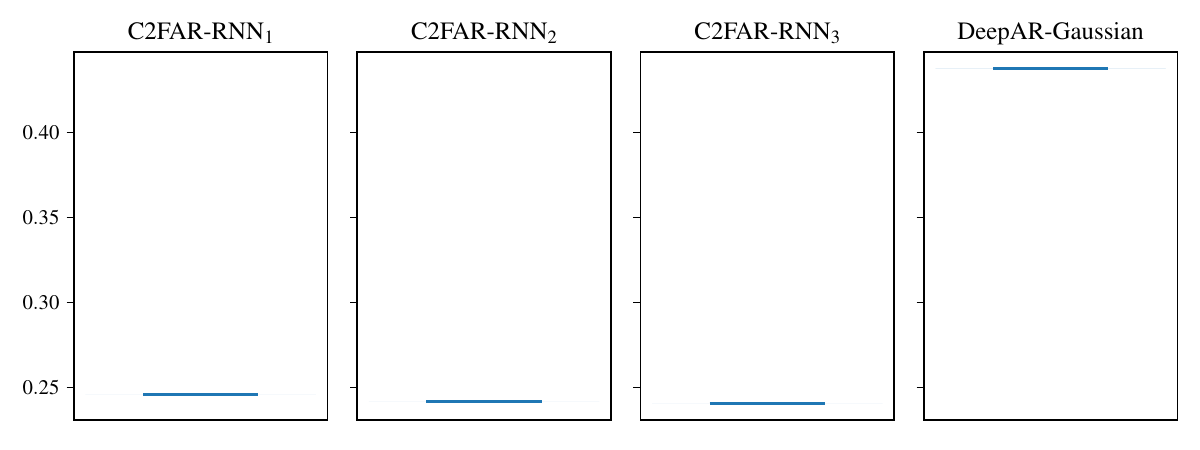}}
      \caption{$\wiki$ sampling stability: distribution of normalized
        deviation (on test set) across different random seeds as a violin plot.\label{fig:sampling_wiki}}
\end{figure}

\begin{figure}
  \centering
      {\includegraphics[width=1.0\textwidth]{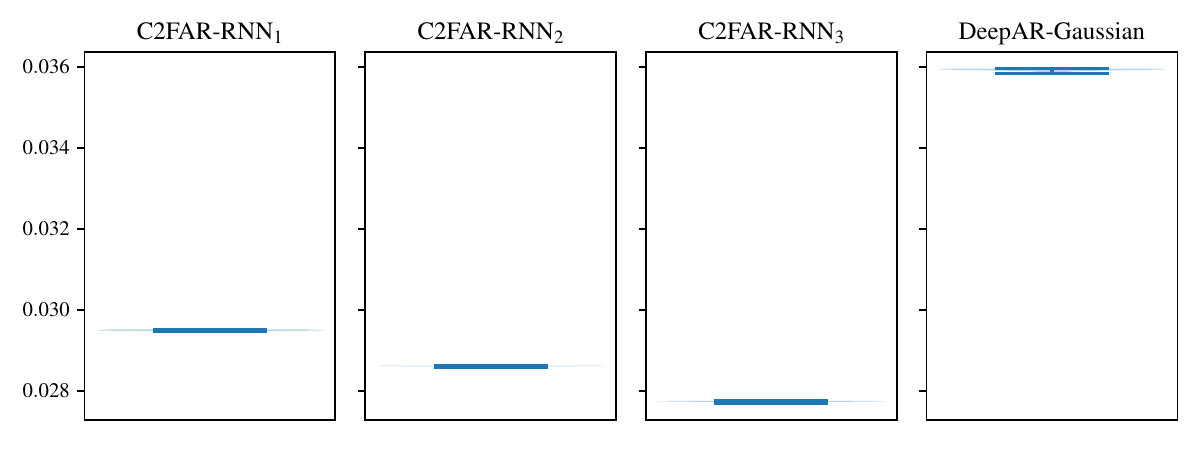}}
      \caption{$\azure$ sampling stability: distribution of normalized
        deviation (on test set) across different random seeds as a violin plot.\label{fig:sampling_azure}}
\end{figure}

Recall that prediction of each example creates a forecast distribution
using 500 different Monte Carlo rollouts of the time series.  We take
the median of this distribution to compute the normalized deviation
(ND).
Overall ND is the average over all test examples, and we report this
in Table~1 of the main paper.
We repeated the ND evaluation of our models 6 times using different
random seeds, and plotted the distribution of the 6 results as
\emph{violin plots} in Figures~\ref{fig:sampling_elec}
to~\ref{fig:sampling_azure}.
The distributions are very narrow, showing that average ND is very
stable with respect to the random seed.
We conclude that evaluation is very stable with respect to the random
seed used in Monte Carlo sampling.

\subsection{Validation/test stability}\label{subsec:valtest}

{\begin{table}
\caption{Validation/test stability\label{tab:devtest}}
\scriptsize
\begin{tabular}{@{}lllllll@{}}
\toprule
     & \multicolumn{3}{c}{$\elec$}
& \multicolumn{3}{c}{$\traffic$}
\\
     & \multicolumn{1}{c}{1st} & \multicolumn{1}{c}{2nd} &
\multicolumn{1}{c}{3rd} & \multicolumn{1}{c}{1st} &
\multicolumn{1}{c}{2nd} & \multicolumn{1}{c}{3rd} \\ \midrule
Val.  & $\ctofartwo$            & \textbf{$\ctofarthree$} &
$\ctofarone$            & $\ctofartwo$            & $\ctofarone$
& $\ctofarthree$          \\
Test & \textbf{$\ctofarthree$} & $\ctofartwo$            &
$\ctofarone$            & $\ctofartwo$            & $\ctofarone$
& $\ctofarthree$          \\ \midrule
     & \multicolumn{3}{c}{$\wiki$}
& \multicolumn{3}{c}{$\azure$}
\\
     & \multicolumn{1}{c}{1st} & \multicolumn{1}{c}{2nd} &
\multicolumn{1}{c}{3rd} & \multicolumn{1}{c}{1st} &
\multicolumn{1}{c}{2nd} & \multicolumn{1}{c}{3rd} \\ \midrule
Val.  & $\ctofartwo$            & $\ctofarone$            &
\textbf{$\ctofarthree$} & $\ctofartwo$            &
\textbf{$\ctofarthree$} & $\ctofarone$            \\
Test & \textbf{$\ctofarthree$} & $\ctofartwo$            &
$\ctofarone$            & \textbf{$\ctofarthree$} & $\ctofartwo$
& $\ctofarone$           
\\ \bottomrule
\end{tabular}
\end{table}
}

Another possible source of instability is different behavior of
systems on the validation set versus the test set.
We summarize this behavior in our data by showing the ranking of the
systems on the validation set and test set in Table~\ref{tab:devtest}
(note we exclude $\deepargaussian$ from this table as it is always
ranked 4th\@ on all validation and test sets).
Results are fairly stable moving from validation to test.
On all validation sets and all test sets, all $\ctofar$ models improve
over $\deepargaussian$.
Also, on all validation sets and all test sets, $\ctofartwo$ improves
over $\ctofarone$.
The ranking in $\traffic$ is the same on validation and test set, but
on the others, the only difference is $\ctofarthree$ moved to first
place on test.

\subsection{Tuning stability}\label{subsec:tuning}

\begin{figure}
  \centering
      {\includegraphics[width=1.0\textwidth]{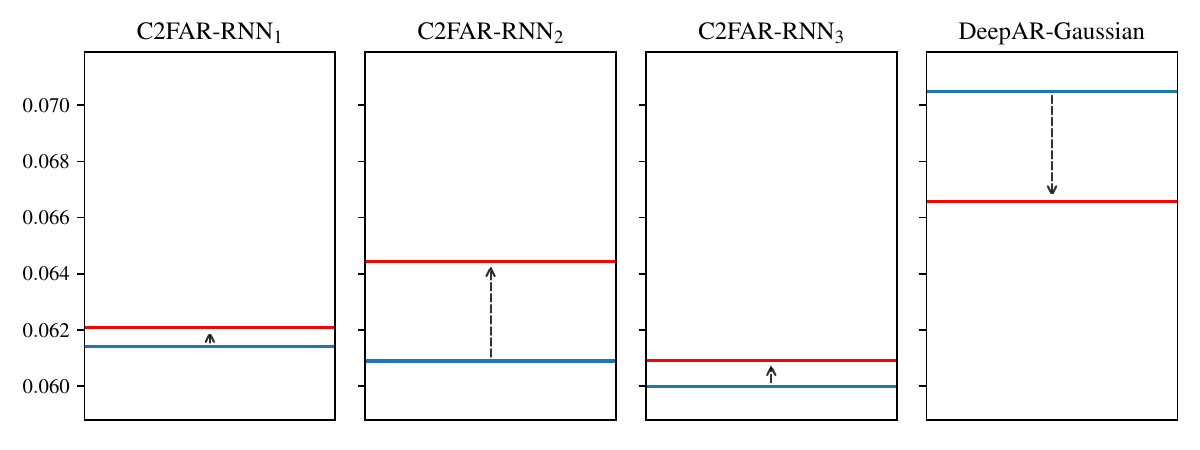}}
      \caption{$\elec$ tuning stability: difference in normalized
        deviation (on test set) from top-1 tuned model (in blue)
        to second-best tuned model (in red).\label{fig:devtest_elec}}
\end{figure}

\begin{figure}
  \centering
      {\includegraphics[width=1.0\textwidth]{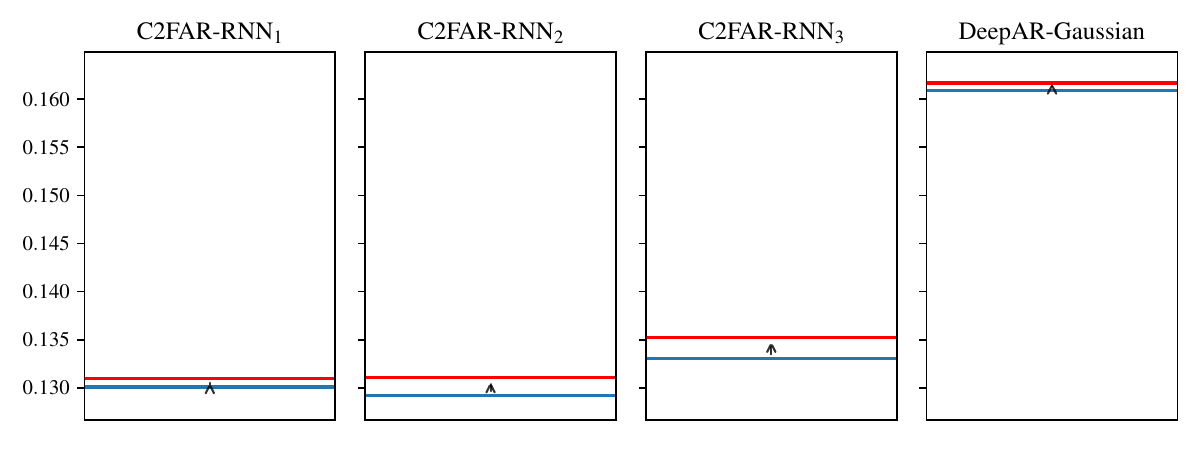}}
      \caption{$\traffic$ tuning stability: difference in normalized
        deviation (on test set) from top-1 tuned model (in blue)
        to second-best tuned model (in red).\label{fig:devtest_traffic}}
\end{figure}

\begin{figure}
  \centering
      {\includegraphics[width=1.0\textwidth]{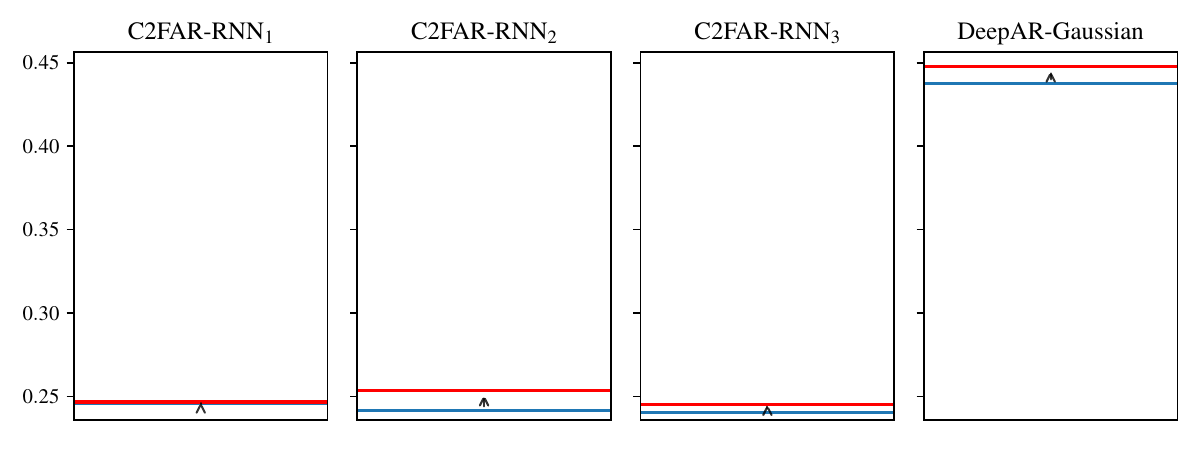}}
      \caption{$\wiki$ tuning stability: difference in normalized
        deviation (on test set) from top-1 tuned model (in blue)
        to second-best tuned model (in red).\label{fig:devtest_wiki}}
\end{figure}

\begin{figure}
  \centering
      {\includegraphics[width=1.0\textwidth]{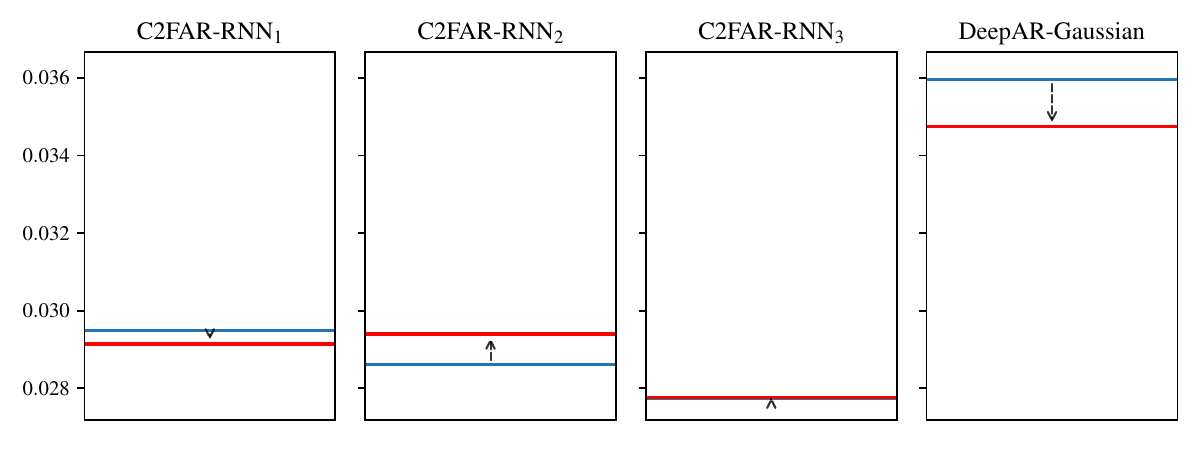}}
      \caption{$\azure$ tuning stability: difference in normalized
        deviation (on test set) from top-1 tuned model (in blue)
        to second-best tuned model (in red).\label{fig:devtest_azure}}
\end{figure}

We assess the stability of our tuning results by considering the
consequences had the optimizer not found the top model on the
validation set.
We therefore evaluate the second-best models from the validation set
on the test set.
We hypothesize that the second-best models may have larger drops for
$\ctofartwo$ and $\ctofarthree$, as these models have more
hyperparameters and are therefore more vulnerable to sub-optimal
tuning.
Figures~\ref{fig:devtest_elec} to~\ref{fig:devtest_azure} show the
results, with the blue lines indicating the original top-model result,
and the red lines indicating the result after switching to the
second-best validation model.

We do see larger drops for the multi-level $\ctofar$ models,
especially $\ctofartwo$, which drops significantly on the $\elec$
dataset.
However, even with the second-best models, $\ctofar$ models perform
better than $\deepargaussian$ across all test sets, and $\ctofarthree$
remains, as before, the top system on three of the four test sets
(\S\ref{subsec:tuning}).
Overall, this suggests that we may wish to use more tuning trials for
the multi-level $\ctofar$ models, or, as suggested in the main paper,
simply use the same number of bins at each level, reducing the number
of bins to a single hyperparameter, as in the flat binning.

\section{Test-set likelihood experiments}\label{sec:likelihoodexp}

{\begin{table}
\caption{Tuning results, tuning for NLL, on $\azure$ with noise added.
  Compare to Table~\ref{tab:tuning_results} for tuning for
  ND\@.\label{tab:nll_tuning}} \footnotesize
\begin{tabular}{llllllll}
\toprule
Dataset  & System            & nhidden & NBins1 & NBins2 & NBins3 &
Total bins & Total intervals \\ \midrule
$\azure$ & $\deepargaussian$ & 74      & -      & -      & -      & -
& -               \\
$\azure$ & $\ctofarone$      & 165     & 638    & -      & -      &
638        & 638             \\
$\azure$ & $\ctofartwo$      & 126     & 70     & 31     & -      &
101        & 2170            \\
$\azure$ & $\ctofarthree$    & 141     & 6      & 14     & 67     & 87
& 5628           \\ \bottomrule
\end{tabular}
\end{table}
}

In this section, we report some supplementary results comparing our
systems for their ability to estimate the log-likelihood of held-out
test data.
As mentioned in the main paper in \S4.2, log-likelihood is sometimes
regarded as the de facto standard for evaluating generative
models~\cite{theis2015note}, and results on log-likelihood estimation
are regarded as a proxy for the effectiveness of models on other tasks
such as anomaly detection or missing value imputation.
Since we cannot \emph{tune} $\ctofar$ models directly for
log-likelihood on discrete data (as this leads to narrower and
narrower density spikes), we investigate models trained under two other
regimes:
\begin{enumerate}
\item Models tuned for ND (that is, the same models used in the
  forecasting experiments)
\item Models tuned for negative log likelihood (NLL), but on
  \emph{dequantized} data, i.e., data with \verb+Uniform[0,1]+ noise
  added, as in prior work~\cite{rasul2020multi}
\end{enumerate}
For the models tuned for NLL on the noise-added data, we use the same
tuning setup as in forecasting, doing 100 tuning evaluations for each
system and using the same tuned and fixed hyperparameters as described
in \S\ref{sec:tuning_setup}.  We performed this experiment on $\azure$
data only.

The resulting tuned hyperparameters are given in
Table~\ref{tab:nll_tuning}.  Interestingly, the tuner selects quite
many more bins for the $\ctofarone$ model as were selected when tuning
for ND (Table~\ref{tab:tuning_results}), suggesting precision is
important even in noise-added data.

{\begin{table}
\caption{NLL results on noisy and original datasets\label{tab:nll}}
\footnotesize
\begin{tabular}{@{}llllllllll@{}}
\toprule
\multirow{3}{*}{Dataset} & \multicolumn{5}{c}{Experimental
  configuration} & \multicolumn{4}{c}{System}
\\
 &
  \begin{tabular}[c]{@{}l@{}}Trained\\ for\end{tabular} &
  \begin{tabular}[c]{@{}l@{}}Tested\\ for\end{tabular} &
  \begin{tabular}[c]{@{}l@{}}Tuned\\ for\end{tabular} &
  \begin{tabular}[c]{@{}l@{}}Trained\\ on\end{tabular} &
  \begin{tabular}[c]{@{}l@{}}Tested\\ on\end{tabular} &
  \begin{tabular}[c]{@{}l@{}}DeepAR-\\ Gaussian\end{tabular} &
  \begin{tabular}[c]{@{}l@{}}C2FAR-\\ RNN$_1$\end{tabular} &
  \begin{tabular}[c]{@{}l@{}}C2FAR-\\ RNN$_2$\end{tabular} &
  \begin{tabular}[c]{@{}l@{}}C2FAR-\\ RNN$_3$\end{tabular} \\ \midrule
$\azure$                 & NLL    & NLL   & NLL   & +Noise     &
  +Noise     & 1.355 & -2.011 & \textbf{-2.075} & \textbf{-2.075} \\
$\azure$                 & NLL    & NLL   & ND   & Original   &
  +Noise     & 0.766 & -1.739 & -1.574          & -1.504          \\ \midrule
$\azure$                 & NLL    & NLL   & NLL   & +Noise     &
  Original   & 2.307 & -3.300 & -4.110          & -4.043          \\
$\azure$                 & NLL    & NLL   & ND   & Original   &
  Original   & 0.094 & -2.533 & -4.475          & \textbf{-6.309}
  \\ \bottomrule
\end{tabular}
\end{table}

}

We evaluated both the NLL- and ND-tuned systems on both the original
test data and the test data with \verb+Uniform[0,1]+ noise added.
Results are given in Table~\ref{tab:nll}.  Likelihood is computed in
the normalized domain (after min-max scaling) for all time series.

We can divide the evaluation into two objectives: NLL on the
noise-added data, and NLL on the original data.  If our objective is
NLL on the noise-added data, then we see that multi-level $\ctofar$
models still offer benefits over a flat binning.  One may have
expected this to not be the case, as adding noise removes some of the
precision in the data and thus one of the advantages of $\ctofar$, but
we see that both $\ctofartwo$ and $\ctofarthree$ still prove superior
to the flat binning in this case.  Even on noise-added data,
multi-level $\ctofar$ models use many more total intervals than those
used by a flat binning (Table~\ref{tab:nll_tuning}), illustrating that
precision is still important even in noise-added data.

Now, regarding NLL of the \emph{original} data, we see that
multi-level $\ctofar$ models are again superior to flat binning, and
moreover, multi-level $\ctofar$ models trained for ND are superior to
those trained for NLL on the noise-added data.
This illustrates that, if our objective is to achieve maximum
likelihood on the original data, adding noise to dequantize the data
may not be a good solution; here it results in worse NLL than simply
tuning for ND\@ on the original data.
We repeat the point in the main paper that, in reality, for tasks such
as anomaly detection, missing value imputation, denoising,
compression, etc., we should not tune for NLL at all, but rather tune
for the application-specific metric of interest.

\end{document}